\documentclass[11pt]{article}

\usepackage[preprint]{acl}
\usepackage{times}
\usepackage{latexsym}
\usepackage{algorithm}
\usepackage{algpseudocode}

\usepackage[T1]{fontenc}
\usepackage[utf8]{inputenc}

\usepackage{microtype}

\usepackage{inconsolata}
\usepackage{booktabs}
\usepackage{multirow}
\usepackage{graphicx}
\usepackage{amsmath}
\usepackage{amssymb}
\newtheorem{proposition}{Proposition}

\usepackage{subcaption}
\title{Deconstructing Off-Policy Ratios: Entropy-Normalized Trust Regions for Asynchronous Reinforcement Learning
}

\author{
\textbf{Guanqun Zhao}$^{\ast\,1,5}$,\;
\textbf{Zijun Xie}$^{\ast\,2,5}$,\;
\textbf{Binbin Zheng}$^{\ast\,3,5}$,\;
\textbf{Yehan Yang}$^{4}$ \\
\textbf{Jiafeng Lu}$^{5}$,\;
\textbf{Aoqi Hu}$^{5}$,\;
\textbf{Enlei Gong}$^{\dagger\,5}$,\;
\textbf{Zeyu Chen}$^{\dagger\,5}$ \\
$^1$Beijing University of Posts and Telecommunications \quad
$^2$Peking University \\
$^3$University of Science and Technology of China \\
$^4$Institute of Computing Technology, Chinese Academy of Sciences \quad
$^5$Baidu Inc. \\
\texttt{zhao-guanqun@bupt.edu.cn} \quad
\texttt{\{lujiafeng,\,gongenlei,\,chenzeyu01\}@baidu.com} \\
\smallskip
{\small $^\ast$Work done during an internship at Baidu Inc.\quad
$^\dagger$Corresponding authors.}
}

\begin{document}
\maketitle
\begin{abstract}
Asynchronous reinforcement learning (RL) accelerates large language model (LLM)
post-training by overlapping rollout generation with policy optimization, but the
resulting stale, off-policy data destabilizes optimization and can cause policy
collapse. Existing methods gate tokens by ratio magnitude alone, applying one
threshold at every position. We show that the ratio's natural scale is set by
token entropy, so deviations from mid-trajectory weight updates stay within this
scale and carry genuine exploration. We further identify an overlooked
low-entropy regime that breaks this scaling, where a near-zero probability
amplifies train--inference mismatch into noise far beyond what the local entropy
admits. A magnitude threshold admits this noise and discards the exploration.
We therefore propose the Entropy-Normalized Trust Region (ENTR). Across
long-horizon agentic tasks and mathematical reasoning benchmarks, ENTR
outperforms existing asynchronous methods. It improves avg@1 on BrowseComp-Plus by $6.9\%$ over the strongest
baseline, trains stably up to $30$ policy versions of staleness, and matches
synchronous GRPO at a $2.6\times$ speedup.
\end{abstract}

%==============================================================
\section{Introduction}
\label{sec:intro}
%==============================================================

\begin{figure}[t]
\centering
\includegraphics[width=0.99\columnwidth]{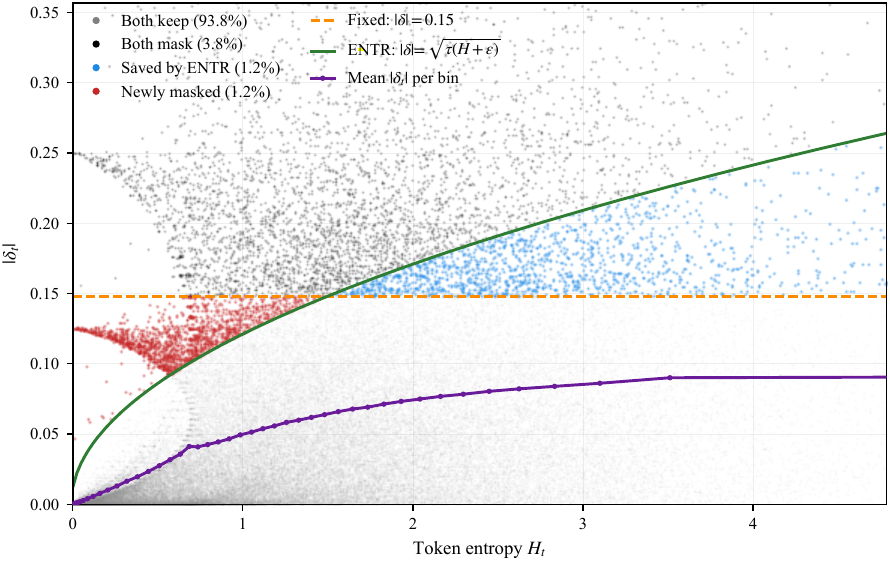}
\caption{Per-bin mean $|\delta_t|$ (purple) grows with entropy $H_t$. A fixed
threshold (dashed) confuses low-entropy noise (red) with high-entropy
exploration (blue); our entropy-normalized boundary (solid) separates them.}
\label{fig:joint}
\end{figure}

\begin{figure*}[t]
\centering
\includegraphics[width=0.99\textwidth]{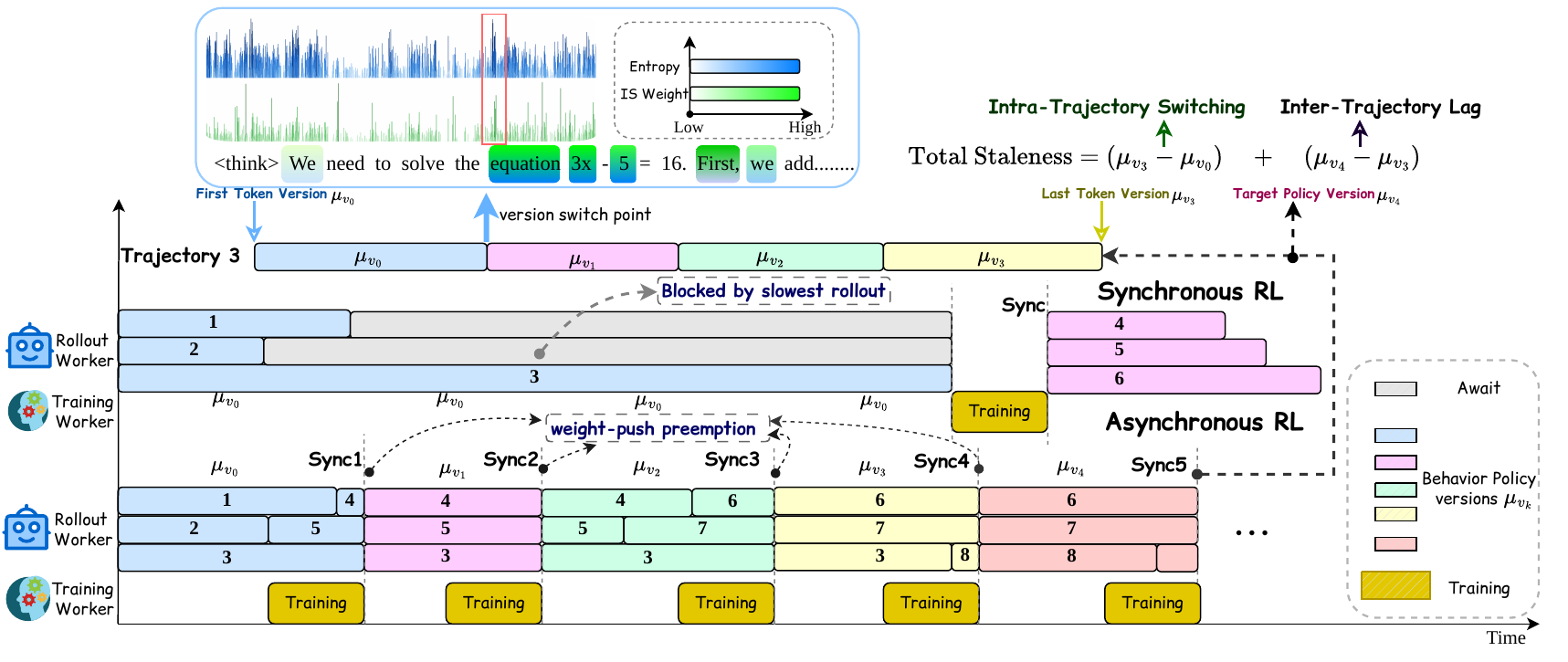}
\caption{Synchronous vs. asynchronous agentic RL. Asynchronous RL removes the
idle pipeline bubbles of synchronous RL, at the cost of a single trajectory
spanning multiple weight versions $\mu_{v_t}$. The resulting staleness decomposes
into \emph{intra-trajectory version switching} and \emph{inter-trajectory lag},
with the former inducing abrupt surges in token entropy and importance ratios at switch points.}
\label{fig:async}
\end{figure*}

Reinforcement learning (RL) is now the dominant paradigm for aligning large
language models~\cite{deepseekai2026deepseekv4,shao2024deepseekmathpushinglimitsmathematical,yu2025dapoopensourcellmreinforcement,zheng2025groupsequencepolicyoptimization} and for eliciting long-horizon
reasoning~\cite{chen2025browsecompplusfairtransparentevaluation,yu2025dapoopensourcellmreinforcement,zheng2025groupsequencepolicyoptimization}
and agentic behavior~\cite{glm5team2026glm5vibecodingagentic}. To scale RL to
long, multi-turn rollouts, modern systems increasingly adopt asynchronous
training~\cite{Sheng_2025,lu2025iirollflash,piché2025pipelinerlfasteronpolicyreinforcement,hu2026dorascalableasynchronousreinforcement,wu2025llamarldistributedasynchronousreinforcement,Bai_2026,li2026staleflowstalenessawaredatamanagement,radixark2026milesv01productionlevelposttraining,wang2026r2podecouplingrolloutinference,tan2026bidirectionalresourceschedulingdisaggregated,bai2026asyncwebrlefficientasynchronousreinforcement,han2026asyncflowasynchronousstreamingrl,lu2026accerldistributedasynchronousreinforcement}.
This design dramatically improves
throughput~\cite{li2026unleashingefficientasynchronousrl} but breaks the
on-policy assumption~\cite{li2026trustregionmaskinglonghorizon,yao2025offpolicy,qi2025precisionrl,ma2025stabilizingmoereinforcementlearning}. Each update
consumes rollouts from behavior policies that lag behind and diverge from the
target policy, and a single long trajectory may even span several weight
versions~\cite{hou2026singlerolloutasynchronousoptimizationagentic}. Optimizing against such stale, ill-defined behavior policies injects
high-variance off-policy gradients that can drive training into irreversible
policy collapse~\cite{zheng2025prosperitycollapsefaroffpolicy}.
Mitigating this instability without sacrificing these gains is thus a central
challenge.

Existing stabilization methods act on the importance ratio in different ways.
Interval clipping bounds it within a fixed
range~\cite{li2026bandpobridgingtrustregions,cagatan2026clippingfree,fu2026logpipitaming,luo2026ratiovarianceregularizedpolicyoptimization,xie2026acpoadaptivecreditpolicy},
hard masking discards the tokens that leave
one~\cite{yao2025offpolicy,lingteam2025stepevolvesscalingreinforcement,KPop2026,zheng2025prosperitycollapsefaroffpolicy,feng2025groupingrouppolicyoptimizationllm,xu2025singlestreampolicyoptimization,wang2026capocriticguidedactionalignedpolicy,ma2026fipoelicitingdeepreasoning},
staleness-mismatch decoupling removes the lag at the systems
level~\cite{fu2026areallargescaleasynchronousreinforcement,guan2026missingoldlogitsasynchronous},
and off-policy objective design replaces the estimator
itself~\cite{ritter2026llmslearnreasonoffpolicy,yuan2025trajectorybellmanresidualminimization}.
In all of them the keep-or-drop decision is read from the ratio's
magnitude, under a single threshold shared by every token position.
We find that the natural scale of a token's deviation is set by its entropy.
A single threshold therefore admits
noise at low entropy and discards genuine exploration at high entropy. Nor is the reference of that ratio well defined, since each long trajectory is
produced by a mixture of weight versions at varying distance from the target
rather than by a single behavior policy.

Motivated by these empirical findings, we propose the Entropy-Normalized Trust Region (ENTR). Specifically, our main contributions are:

\begin{itemize}
    \item We find that off-policy deviations typically grow with token
    entropy in real asynchronous training trajectories. Decomposing
    asynchronous staleness, we further observe that abrupt ratio deviations
    at intra-trajectory version switches rise together with entropy,
    suggesting that these deviations may reflect increased exploration
    rather than corruption.

    \item We identify an overlooked low-entropy regime where deviations
depart from this trend and become unusually large relative to local
entropy. For low-probability tokens sampled at these positions, even
small probability discrepancies can translate into large log-ratio
deviations.

   \item We propose the Entropy-Normalized Trust Region (ENTR), which
suppresses low-entropy outliers while allowing larger high-entropy
deviations. ENTR improves BrowseComp-Plus avg@1 by $6.9\%$ over
the strongest asynchronous baseline, trains stably with $30$ steps
of off-policy lag, and approaches synchronous GRPO performance
with a $2.6\times$ per-step speedup.

\end{itemize}

%==============================================================
\section{Related Work}
\label{sec:related}
%==============================================================
To mitigate the instability of asynchronous LLM RL, prior works have explored various stabilization strategies.

\subsection{Importance-Sampling Corrections}
A prominent line of work controls variance by intervening directly on the importance ratio. One family discards out-of-bounds tokens directly, either by interval clipping~\cite{fu2026logpipitaming,qi2026rethinkingtrustregionllm,shen2026vespovariationalsequencelevelsoft,xie2026acpoadaptivecreditpolicy,yang2026stalestablestalenessadaptivetrust,li2026trustregionmaskinglonghorizon,li2026bandpobridgingtrustregions,sheng2026espoentropyimportancesampling,huang2026stableasynchronyvariancecontrolledoffpolicy,min2026orchestratingtokenssequencesdynamic}, which constrains each ratio to a fixed interval, or by hard
masking~\cite{huang2026stableasynchronyvariancecontrolledoffpolicy,li2026trustregionmaskinglonghorizon,liu2026asympoasymmetricscalepolicyoptimization},
which zeroes out the contribution of any token whose deviation exceeds a preset
bound. IcePop~\cite{lingteam2025stepevolvesscalingreinforcement} thresholds the
ratio directly, whereas KPop~\cite{KPop2026} thresholds a bidirectional binary
KL between the behavior and target token probabilities, and
M2PO~\cite{zheng2025prosperitycollapsefaroffpolicy} masks the largest squared
log-ratios until their batch mean falls below a fixed budget.

\subsection{Staleness--Mismatch Decoupling and Off-Policy Objective Design}
Staleness--mismatch
decoupling~\cite{fu2026areallargescaleasynchronousreinforcement,guan2026missingoldlogitsasynchronous}
approximates the behavior policy that generated each rollout, separating
genuine policy drift from benign staleness so that only the former is
corrected. This presumes a single behavior policy per trajectory. In long
multi-turn rollouts, in-flight weight updates switch the inference engine
mid-generation, so a trajectory is produced by a mixture of weight
versions and its behavior policy is not recoverable. Off-policy objective
design~\cite{ritter2026llmslearnreasonoffpolicy,shen2026vespovariationalsequencelevelsoft,luo2026ratiovarianceregularizedpolicyoptimization,yuan2025trajectorybellmanresidualminimization}
instead reuses stale data without hard cutoffs, softening clipping into a
distributional variance penalty or regressing against a reference policy
without importance ratios. Yet the former imposes a batch-level,
position-invariant budget, and the latter forgoes importance correction
altogether. Both families therefore treat every token alike, with no account of
the entropy that sets its natural scale.

Across these lines, the boundary of trust is fixed either by ratio magnitude or
at the sequence level, decoupled from the local state of each token. ENTR
instead scales the trust region to each token's entropy, the natural scale of
its off-policy deviation.

\section{Motivation}
\label{sec:motivation}

To understand why magnitude-only suppression struggles in asynchronous agentic RL, we analyze the joint distribution of token entropy $H_t$ and the log-importance-ratio $|\delta_t|$ on real rollouts using Qwen3-30B-A3B on BrowseComp-Plus~\cite{chen2025browsecompplusfairtransparentevaluation}. We find that $|\delta_t|$ is inherently coupled with token entropy, suggesting that a single global cutoff may not fully accommodate this state-dependent variance.

\begin{figure}[t]
\centering
\includegraphics[width=0.95\columnwidth]{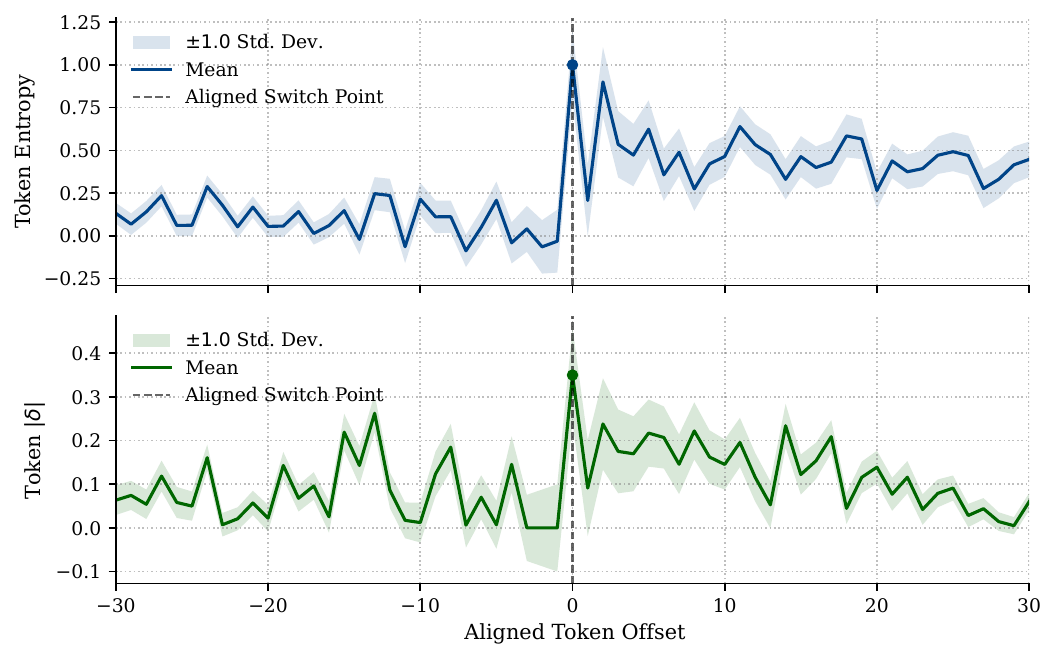}
\caption{Token entropy $H$ and ratio magnitude $|\delta|$ aligned 
at an intra-trajectory version switch where $t=0$. Both surge synchronously.}
\label{fig:version_switch}
\end{figure}

\subsection{Ratio Magnitude Relates to Entropy}
% 这一章对应附录A，也要整理一下，证明要简洁，符号要一致，少用新的符号了
Empirically, the magnitude of the log importance ratio is not uniform
across token positions but generally increases with token entropy, as
shown by the per-bin statistics in Figure~\ref{fig:joint}. Under a local
logit-perturbation model, the conditional second moment of the log-ratio
is governed by the behavior policy's uncertainty and is approximately
proportional to entropy in the bulk regime:
\begin{equation}
\label{eq:law}
\mathbb{E}[\delta_t^2\mid H_t\in\mathcal{B}]
\approx a_t H_t ,
\end{equation}
where $\mathcal{B}$ denotes an entropy bin and $a_t$ absorbs the local
perturbation scale. The assumptions and derivation are provided in Appendix~\ref{app:entropy-ratio}. This observation motivates an entropy-scaled trust region,
which we develop in the next section.

% 这一章对应的附录B，要好好整理，其中的顺序，先讲低熵区，再讲高熵区，
\subsection{Dual Phenomena under Asynchronous Dynamics}

Asynchronous RL removes pipeline bubbles, but weight updates then land
mid-generation, as illustrated in Figure~\ref{fig:async}. Let
$v_{\mathrm{tgt}}$ be the target version at the current update, and
$v_{\mathrm{first}}$, $v_{\mathrm{last}}$ the versions that generated the first
and last tokens of a rollout. We decompose the total staleness
$v_{\mathrm{tgt}}-v_{\mathrm{first}}$ as
\begin{equation}
\label{eq:staleness-decomp}
\Delta^{\mathrm{intra}}\triangleq v_{\mathrm{last}}-v_{\mathrm{first}},
\qquad
\Delta^{\mathrm{inter}}\triangleq v_{\mathrm{tgt}}-v_{\mathrm{last}},
\end{equation}
where $\Delta^{\mathrm{intra}}$ is the \emph{intra-trajectory version switch},
a single long trajectory spanning multiple weight versions.

At such a switch the new weights resume a prefix they did not produce, so the
behavior policy is less committed about how to continue and its entropy rises;
by Eq.~\eqref{eq:law} the natural scale of the deviation rises with it, and
$H_t$ and $|\delta_t|$ surge together at the switch point, as
Figure~\ref{fig:version_switch} shows. The joint surge factorizes into an
entropy factor and a local scale,
\begin{equation}
\label{eq:switch-invariant}
|\delta_t|\approx\sqrt{a_t H_t},
\qquad
z_t\triangleq\frac{\delta_t}{\sqrt{H_t}},
\qquad
\mathbb{E}[z_t^2]\approx a_t .
\end{equation}
The surge is therefore carried entirely by $\sqrt{H_t}$, while the standardized
deviation $z_t$ remains at the local scale $a_t$ and is invariant to it. A fixed
threshold $|\delta_t|\le c$ discards these ratios as drift, although $z_t^2$
stays bounded and they are in fact valid exploration within the natural
high-entropy envelope.

The low-entropy region of Figure~\ref{fig:joint} exposes a largely
overlooked phenomenon that this scaling does not account for. There
$|\delta_t|$ rises far beyond what the local entropy admits, tracing concave
arcs that open as $H_t\to0$. These deviations arise from sampling rather than
from policy drift. At a confident position the model commits to a dominant
token, and the sampler occasionally draws one of the remaining alternatives
instead, whose training-side probability $q_t\triangleq\pi_\theta(y_t\mid s_t)$
is near zero. The entropy then sits at the binary bound
\begin{equation}
\label{eq:hbin}
H_t\approx H_{\mathrm{bin}}(q_t)=-q_t\log q_t-(1-q_t)\log(1-q_t),
\end{equation}
independent of vocabulary size. The rollout and training engines assign
slightly different probabilities to the same token, and this train--inference
mismatch $e_t\triangleq\mu_t(y_t)-q_t$ is zero-mean with
$\operatorname{Var}(e_t\mid q_t)\propto q_t(1-q_t)$. The log-ratio divides
$e_t$ by $q_t$, so a mismatch that is negligible in absolute terms becomes an
unbounded relative one once the denominator vanishes,
\begin{equation}
\label{eq:low-entropy-var}
\mathbb{E}[\delta_t^2\mid q_t]\propto\frac{1-q_t}{q_t}
\xrightarrow[q_t\to0]{}\infty .
\end{equation}
Eqs.~\eqref{eq:hbin} and~\eqref{eq:low-entropy-var} move in opposite
directions as $q_t\to0$, and together they trace the concave arcs.
Appendix~\ref{app:lowentropy-arc} gives the derivation. A fixed-magnitude
threshold admits these outliers, whose heavy-tailed weights $e^{\delta_t}$
carry no learning signal yet dominate the batch gradient. The entropy-scaled
rule contracts around them instead.

\subsection{An Entropy-Scaled Design Principle}

The two regimes share one structure. What separates a valid update from harmful
drift is not the magnitude of $\delta_t$ but its size relative to the scale that
$H_t$ admits at that position. A fixed threshold $|\delta_t|\le c$ imposes a
single boundary on a heteroscedastic quantity and therefore errs in opposite
directions at the two ends, discarding high-entropy exploration while keeping
low-entropy noise, and no choice of $c$ removes both errors at once. Comparing
each deviation against its own scale does. The standardized deviation $z_t$ of
Eq.~\eqref{eq:switch-invariant} is entropy-free wherever the scaling law holds,
so one budget on $z_t^2$ carries the same meaning at every position and widens
the boundary as $\sqrt{H_t}$. Where the law breaks the same normalization acts
in reverse, contracting the boundary as $H_t\to0$ onto exactly those deviations
that diverge. We formalize this next.

\section{Method}
\label{sec:method}
% 这一章对应了CDEF四个附录，太多了太冗长了，后面精简一下，没用的附录直接删掉了，就这样
We now formalize the principle of the preceding section as the
Entropy-Normalized Trust Region (ENTR). We first establish a local
connection between squared log-ratio deviations and KL divergence,
then use behavior-policy entropy to adapt the token-level
acceptance threshold.

At token position $t$ with state $s_t$, let $\mu_t(\cdot \mid s_t)$ denote the
behavior policy that generated token $y_t$, and $\pi_\theta(\cdot \mid s_t)$ the
target policy being optimized. We define the log-importance-ratio $\delta_t$ and
the behavior-policy entropy $H_t$ as:
\begin{align}
    \delta_t &\triangleq \log \frac{\pi_\theta(y_t \mid s_t)}{\mu_t(y_t \mid s_t)}, \label{eq:delta_def} \\
    H_t &\triangleq -\sum_{w \in V} \mu_t(w \mid s_t) \log \mu_t(w \mid s_t). \label{eq:entropy_def}
\end{align}

\subsection{A Trust-Region View of Policy Deviations}

At a fixed state $s_t$, KL divergence provides a distribution-level
measure of the discrepancy between the behavior and target policies.
To connect this quantity to token-level deviations, let
$\delta_t(v)$ denote the log-ratio in Eq.~\eqref{eq:delta_def}
evaluated at any candidate token $v$.
Assuming the two conditional policies have common support,
$\mathbb{E}_{v\sim\mu_t}[\pi_\theta(v)/\mu_t(v)-1]=0$, giving
\begin{equation}
\label{eq:kl2}
D_{\mathrm{KL}}(\mu_t\|\pi_\theta)
=
\mathbb{E}_{v\sim\mu_t}
\left[e^{\delta_t(v)}-1-\delta_t(v)\right].
\end{equation}

For $M>0$, suppose that $|\delta_t(v)|\le M$ for every token
in the support of $\mu_t$.
A pointwise comparison of
$e^{\delta_t(v)}-1-\delta_t(v)$ with $\delta_t(v)^2$
then gives, whenever $\mathbb{E}_{\mu_t}[\delta_t^2]>0$,
\begin{equation}
\label{eq:kl-sandwich}
\frac{D_{\mathrm{KL}}(\mu_t\|\pi_\theta)}
     {\mathbb{E}_{\mu_t}[\delta_t^2]}
\in
\left[
\frac{e^{-M}-1+M}{M^2},
\frac{e^M-1-M}{M^2}
\right].
\end{equation}
Both endpoints approach $\tfrac12$ as $M\to0$.
The proof is given in Appendix~\ref{app:kl_expansion}.

This relation gives the squared log-ratio a local interpretation
in terms of policy divergence, motivating $\delta_t^2$ as a
token-level discrepancy score.
The remaining design choice is how to set its acceptance threshold
across positions with different behavior-policy uncertainty.

\subsection{The Entropy-Normalized Trust Region}
\label{sec:standardized}
The entropy dependence observed in Section~\ref{sec:motivation}
guides this threshold choice. It suggests assessing each deviation
relative to the uncertainty of the behavior policy that generated
the token. We therefore define the normalized discrepancy score
\begin{equation}
\label{eq:score}
S_t\triangleq\frac{\delta_t^2}{H_t+\epsilon},
\end{equation}
where $\epsilon>0$ provides a positive floor for the entropy scale.
The behavior-policy entropy $H_t$ is recorded during rollout
without an additional model forward pass.
A common budget on $S_t$ therefore translates into a
position-dependent bound on the original deviation.

Specifically, for a budget $\tau>0$, a token is accepted when
\begin{equation}
\label{eq:mask}
S_t\le\tau
\quad\Longleftrightarrow\quad
|\delta_t|\le b_{\mathrm{ENTR}}(H_t),
\end{equation}
where
$b_{\mathrm{ENTR}}(H_t)\triangleq\sqrt{\tau(H_t+\epsilon)}$.
Here, the trust region denotes this token-level acceptance region.
It selects sampled-token contributions to the update.
The KL analysis motivates the discrepancy measure,
while entropy sets the acceptance threshold.

The boundary contracts to $\sqrt{\tau\epsilon}$ as $H_t\to0$
and grows approximately as $\sqrt{\tau H_t}$ when
$H_t\gg\epsilon$.
Thus, the same log-ratio deviation can be rejected at a low-entropy
position and accepted at a high-entropy position.
For a fixed reference scale $C>\epsilon$, the constant threshold
$\sqrt{\tau C}$ meets $b_{\mathrm{ENTR}}$ at $H_t=C-\epsilon$.
ENTR is strictly tighter below this entropy level and strictly
wider above it.

\subsection{The ENTR Objective}

\begin{algorithm}[t]
\caption{The ENTR update}
\label{alg:entr}
\begin{algorithmic}[1]
\Require budget $\tau$, entropy floor $\epsilon$, clip range
$(\epsilon_{\mathrm{low}},\epsilon_{\mathrm{high}})$
\While{training}
  \State Draw a prompt batch from $\mathcal{D}$
  \State Sample $G$ responses per prompt under $\mu$
  \State Record $\log\mu_{i,t}(o_{i,t})$ and $H_{i,t}$ from the rollout logits
  \State Compute group-normalized advantages $A_{i,t}$
  \For{each token $(i,t)$}
    \State $\delta_{i,t}\gets\log\pi_\theta(o_{i,t}\mid s_{i,t})-\log\mu_{i,t}(o_{i,t})$
    \State $S_{i,t}\gets\delta_{i,t}^2/(H_{i,t}+\epsilon)$
    \State $M_{i,t}\gets\mathbf{1}[S_{i,t}\le\tau]$
  \EndFor
  \State Update $\theta$ on $\mathcal{L}_{\mathrm{ENTR}}$ of Eq.~\eqref{eq:loss1}
  \State Log $\rho_{\mathrm{mask}}$ of Eq.~\eqref{eq:mask-frac}
  \State Publish $\theta$ to the rollout engine
\EndWhile
\end{algorithmic}
\end{algorithm}

Eq.~\eqref{eq:mask} is a rule on a single token, and we now place it in the
training objective. For the $t$-th token of the $i$-th response in a group of
$G$ responses sampled for one prompt, the importance ratio and its deviation are
\begin{equation}
\label{eq:loss-def}
r_{i,t}=\frac{\pi_\theta(o_{i,t}\mid s_{i,t})}{\mu_{i,t}(o_{i,t})},\qquad
\delta_{i,t}=\log r_{i,t},
\end{equation}
and the keep mask evaluates the score of Eq.~\eqref{eq:score} at that position,
\begin{equation}
\label{eq:mask-def}
M_{i,t}=\mathbf{1}\!\left[S_{i,t}\le\tau\right],
\qquad
S_{i,t}=\frac{\delta_{i,t}^2}{H_{i,t}+\epsilon}.
\end{equation}
The mask and the PPO clip both act on $r_{i,t}$ but do different work. The clip
bounds the step size of tokens the trust region already admits, whereas
$M_{i,t}$ enforces the trust region itself by removing every token that falls
outside $b_{\mathrm{ENTR}}(H_{i,t})$. With group-normalized advantage
$A_{i,t}$, the objective is
\begin{multline}
\label{eq:loss1}
\mathcal{L}_{\mathrm{ENTR}}(\theta)
= -\,\mathbb{E}_{x\sim\mathcal{D},\ \{o_i\}_{i=1}^{G}\sim\mu(\cdot\mid x)} \\
\Bigg[\,
\frac{1}{\sum_{i=1}^{G}|o_i|}\sum_{i=1}^{G}\sum_{t=1}^{|o_i|}
M_{i,t}\cdot \\
\min\!\big(r_{i,t}\,A_{i,t},\ \mathrm{clip}(r_{i,t},\,1-\epsilon_{\mathrm{low}},\,1+\epsilon_{\mathrm{high}})\,A_{i,t}\big)
\Bigg].
\end{multline}

Since $\mu_{i,t}$ is fixed, $\nabla_\theta r_{i,t}=r_{i,t}\nabla_\theta\log\pi_\theta$
and the hard mask is a stop-gradient gate, yielding
\begin{multline}
\label{eq:grad-exact}
\nabla_\theta\mathcal{L}_{\mathrm{ENTR}}
= -\,\mathbb{E}\Bigg[
\frac{1}{\sum_{i}|o_i|}
\sum_{i,t}
M_{i,t}\,\kappa_{i,t}\,r_{i,t}\,A_{i,t}\, \\
\times\,\nabla_\theta\log\pi_\theta(o_{i,t}\mid s_{i,t})
\Bigg],
\end{multline}
where $\kappa_{i,t}\in\{0,1\}$ is the clip indicator derived in
Appendix~\ref{app:grad}. A token with $M_{i,t}=0$ leaves the loss entirely,
whereas a clipped token still enters the loss and contributes no gradient. As a
diagnostic of how aggressively the trust region discards tokens, we track the
masked fraction
\begin{equation}
\label{eq:mask-frac}
\rho_{\mathrm{mask}}
=1-\frac{1}{\sum_{i}|o_i|}\sum_{i,t}M_{i,t}.
\end{equation}

Algorithm~\ref{alg:entr} states the resulting update. The rule adds one
elementwise score and one comparison per token to a standard asynchronous GRPO
step, and $H_{i,t}$ is recorded by the rollout engine from logits it has already
computed.

\begin{table*}[t]
\centering
\small
\setlength{\tabcolsep}{4pt}
\resizebox{\textwidth}{!}{%
\begin{tabular}{lcccccccc}
\toprule
& \multicolumn{7}{c}{Search-R1} & BrowseComp-Plus \\
\cmidrule(lr){2-8}\cmidrule(lr){9-9}
Method & NQ$^\dagger$ & TriviaQA$^\star$ & PopQA$^\star$ & HotpotQA$^\dagger$
& 2Wiki$^\star$ & Musique$^\star$ & Bamboogle$^\star$ & avg@1 \\
\midrule
GRPO(Sync)  & 38.29 & 64.72 & 28.01 & 28.95 & 36.95 & 12.26 & 44.23 & 38.55 \\
GRPO(Async) & 30.43 & 59.79 & 23.20 & 24.07 & 30.14 & 8.52 & 39.97 & 28.91 \\
IcePop      & 32.34 & 61.47 & 24.98 & 27.18 & 30.69 & 9.58 & 40.10 & 32.53 \\
KPop        & 32.17 & 61.53 & 25.55 & 27.05 & 32.50 & 10.78 & 41.61 & 34.94 \\
M2PO        & 36.05 & 62.15 & 26.19 & 28.17 & 34.58 & 10.50 & 40.89 & 33.72 \\
\textit{ENTR (Ours)}
 & \textbf{\textit{38.37}} & \textbf{\textit{64.41}} & \textbf{\textit{27.34}}
 & \textbf{\textit{28.51}} & \textbf{\textit{36.72}} & \textbf{\textit{11.55}}
 & \textbf{\textit{43.31}} & \textbf{\textit{37.34}} \\
\midrule
& \multicolumn{2}{c}{AIME24} & \multicolumn{2}{c}{AIME25}
& \multicolumn{2}{c}{AIME26} & \multicolumn{2}{c}{GSM8K-Tool} \\
\cmidrule(lr){2-3}\cmidrule(lr){4-5}\cmidrule(lr){6-7}\cmidrule(lr){8-9}
Method & avg@4 & pass@4 & avg@4 & pass@4 & avg@4 & pass@4
& avg@4 & pass@4 \\
\midrule
GRPO(Sync)  & 19.84 & 27.73 & 16.67 & 27.20 & 16.12 & 28.12 & 96.07 & 98.49 \\
GRPO(Async) & 14.17 & 25.69 & 14.16 & 22.57 & 12.50 & 20.75 & 60.72 & 88.72 \\
IcePop      & 17.52 & 27.42 & 15.78 & 25.45 & 14.16 & 22.01 & 65.01 & 93.35 \\
KPop        & 18.74 & 26.61 & 15.12 & 25.08 & 14.79 & 24.73 & 70.51 & 93.28 \\
M2PO        & 18.53 & 28.08 & \textbf{16.03} & 24.97 & 14.32 & 26.75 & 95.28 & 97.76 \\
\textit{ENTR (Ours)} & \textbf{\textit{20.03}} & \textbf{\textit{31.78}}
 & \textit{15.64} & \textbf{\textit{26.23}}
 & \textbf{\textit{15.46}} & \textbf{\textit{27.14}}
 & \textbf{\textit{95.69}} & \textbf{\textit{98.87}} \\
\bottomrule
\end{tabular}}
\caption{Main results. ENTR attains the best asynchronous score on fifteen of
the sixteen columns and closes most of the gap to synchronous GRPO.
$\dagger$ and $\star$ mark in-domain and out-of-domain evaluation sets for
Search-R1.}

\label{tab:main}
\end{table*}

\section{Experiment}
\subsection{Setup}

\begin{figure*}[t]
\centering
\begin{subfigure}[t]{0.24\textwidth}
\centering
\includegraphics[width=\linewidth]{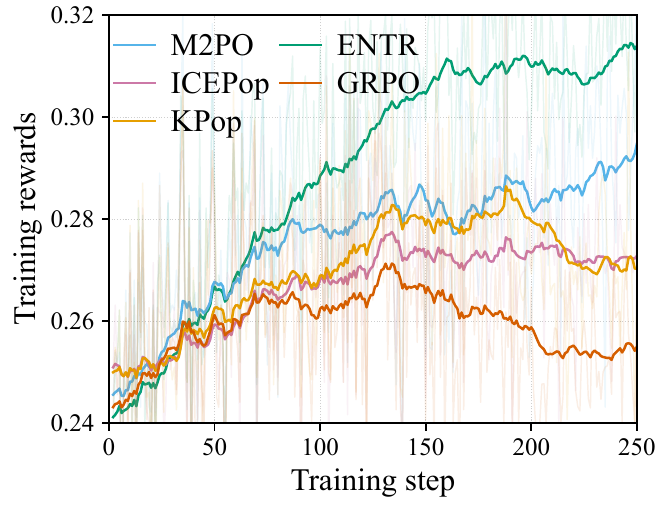}
\caption{Search-R1}
\label{fig:searchr1_score}
\end{subfigure}
\hfill
\begin{subfigure}[t]{0.24\textwidth}
\centering
\includegraphics[width=\linewidth]{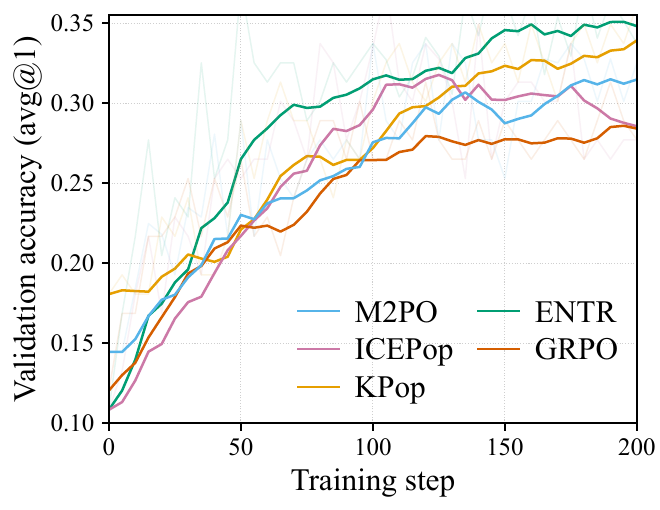}
\caption{BrowseComp-Plus}
\label{fig:val_acc}
\end{subfigure}
\hfill
\begin{subfigure}[t]{0.24\textwidth}
\centering
\includegraphics[width=\linewidth]{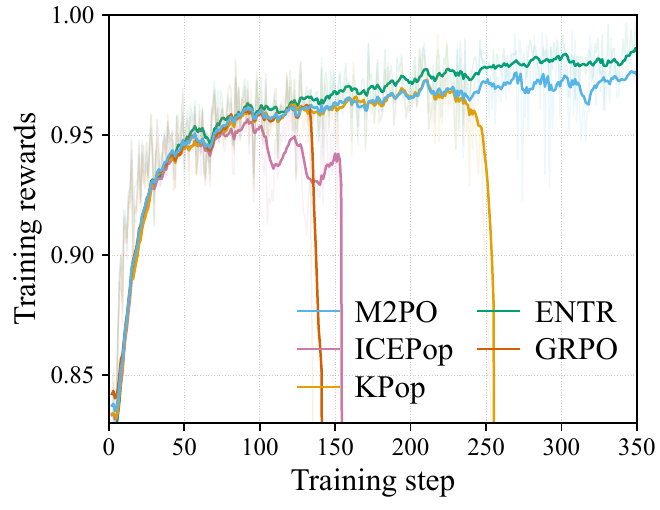}
\caption{GSM8K}
\label{fig:gsm8k_score}
\end{subfigure}
\hfill
\begin{subfigure}[t]{0.24\textwidth}
\centering
\includegraphics[width=\linewidth]{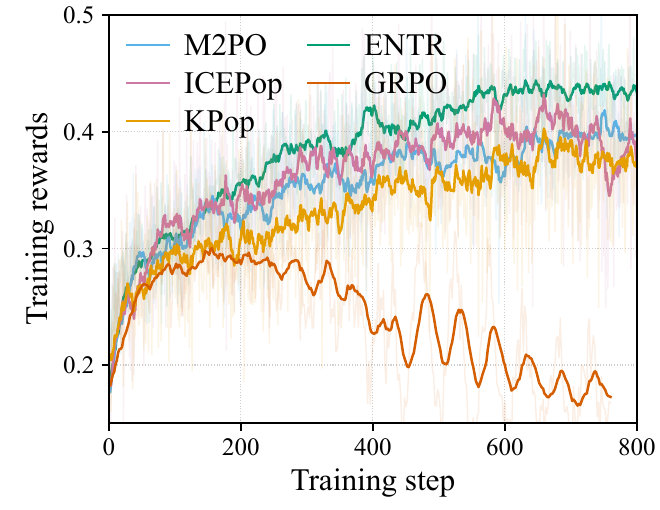}
\caption{DAPO-Math}
\label{fig:dapo_score}
\end{subfigure}
\caption{Training curves on four benchmarks under asynchronous rollout. ENTR
improves throughout on every benchmark, while GRPO, ICEPop and KPop either
collapse or plateau.}
\label{fig:main_curves}
\end{figure*}

\paragraph{Tasks.}
We evaluate ENTR on two task families. The first is multi-turn agentic
tool use, where a rollout is long and interleaves reasoning with external tool
calls. It covers agentic search on
Search-R1~\cite{jin2025searchr1trainingllmsreason} with Qwen3-8B~\cite{qwen3technicalreport},
agentic deep search on
BrowseComp-Plus~\cite{chen2025browsecompplusfairtransparentevaluation} with
Qwen3-30B-A3B~\cite{qwen3technicalreport}, and tool-integrated math on
multi-turn GSM8K~\cite{cobbe2021gsm8k} with Qwen2.5-7B~\cite{qwen2.5} under a
binary exact-match reward. The second is single-turn mathematical
reasoning, which tests generalization beyond the long-horizon regime. We train
on DAPO-Math~\cite{yu2025dapoopensourcellmreinforcement} with
Qwen2.5-7B~\cite{qwen2.5} and evaluate on AIME~2024~\cite{aime24},
2025~\cite{aime25}, and 2026~\cite{aime26}.

\paragraph{Baselines.}
We compare against asynchronous methods that apply a token-level accept rule on
the same ratio $\pi_\theta/\mu$ as ENTR. All runs share an identical
experimental configuration, isolating this rule as the only difference.
Sync is synchronous GRPO~\cite{Guo_2025} and serves as an upper reference.
Async is vanilla asynchronous training.
IcePop~\cite{lingteam2025stepevolvesscalingreinforcement} keeps a token when its
ratio lies within a fixed interval.
KPop~\cite{KPop2026} keeps a token when the bidirectional binary KL between the
behavior and target token probabilities stays below a threshold.
M2PO~\cite{zheng2025prosperitycollapsefaroffpolicy} masks the largest squared
log-ratios until their batch mean falls below a fixed budget.

\paragraph{Implementation.}
All experiments are implemented in the verl framework~\cite{Sheng_2025} on
nodes of eight H800 GPUs, using disaggregated rollout and training pools in the
asynchronous setting. Every baseline is run under the best
configuration reported in its original paper. Full hyperparameters, per-task
resource splits and training configurations are given in
Appendix~\ref{app:training-config}.

\begin{figure*}[t]
\centering
\begin{subfigure}[b]{0.32\textwidth}
\centering
\includegraphics[width=\linewidth]{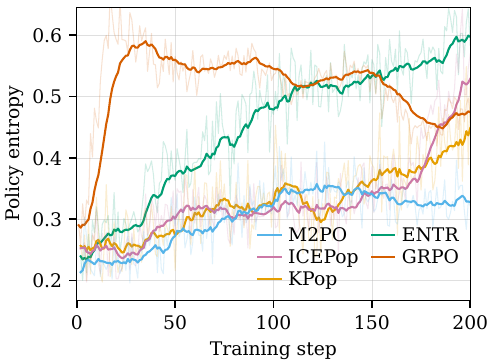}
\caption{Policy entropy on BrowseComp-Plus}
\label{fig:bcp_entropy}
\end{subfigure}
\hfill
\begin{subfigure}[b]{0.32\textwidth}
\centering
\includegraphics[width=\linewidth]{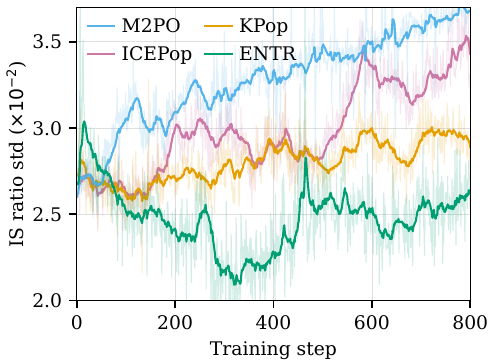}
\caption{IS-ratio deviation on DAPO-Math}
\label{fig:dapo_is_std}
\end{subfigure}
\hfill
\begin{subfigure}[b]{0.32\textwidth}
\centering
\includegraphics[width=\linewidth]{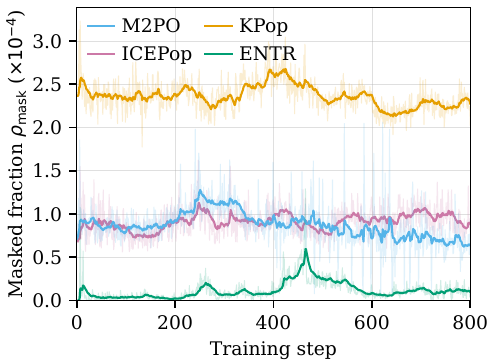}
\caption{Masked fraction on DAPO-Math}
\label{fig:dapo_mask}
\end{subfigure}
\\[2mm]
\begin{subfigure}[b]{0.32\textwidth}
\centering
\includegraphics[width=\linewidth]{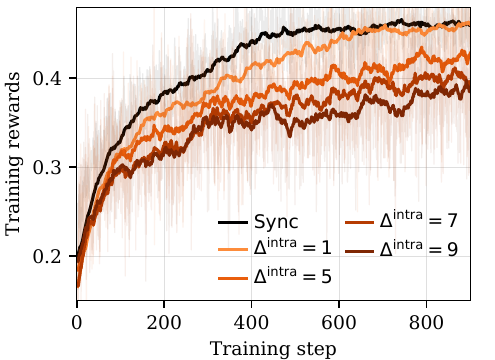}
\caption{Intra-trajectory staleness sweep}
\label{fig:intra_sweep}
\end{subfigure}
\hspace{0.02\textwidth}
\begin{subfigure}[b]{0.32\textwidth}
\centering
\includegraphics[width=\linewidth]{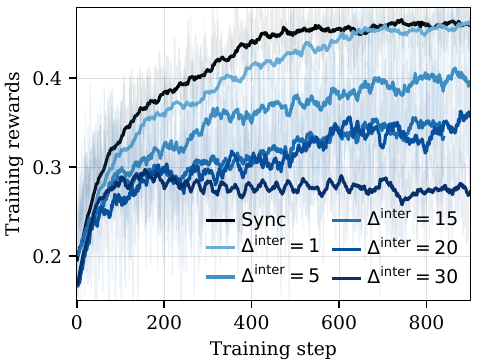}
\caption{Inter-trajectory staleness sweep}
\label{fig:inter_sweep}
\end{subfigure}
\caption{Training dynamics and staleness sweeps. Top, policy entropy on
BrowseComp-Plus and the masked fraction and IS-ratio deviation on DAPO-Math.
Bottom, each staleness source swept in isolation on DAPO-Math.}
\label{fig:dynamics}
\end{figure*}

\subsection{Long-Horizon Multi-Turn Tool Use}

The three agentic settings distribute the staleness of
Eq.~\eqref{eq:staleness-decomp} differently, with
$\Delta^{\mathrm{intra}}{=}2$ and $\Delta^{\mathrm{inter}}{=}9$ on Search-R1,
$\Delta^{\mathrm{intra}}{=}5$ and $\Delta^{\mathrm{inter}}{=}1$ on
BrowseComp-Plus, and $\Delta^{\mathrm{intra}}{=}5$ and
$\Delta^{\mathrm{inter}}{=}13$ on tool-integrated GSM8K.

On Search-R1, ENTR terminates above every baseline in the training reward of
Figure~\ref{fig:searchr1_score}, exceeding M2PO by $6.5\%$, IcePop by
$15.1\%$, KPop by $16.2\%$ and naive asynchronous GRPO by $23.1\%$ in relative
terms. Averaged over the seven evaluation sets of Table~\ref{tab:main} it
reaches $35.74$ against $34.08$ for M2PO, the strongest baseline, and recovers
all but $0.46$ of the $5.33$ points that asynchrony costs the synchronous run.
On BrowseComp-Plus, the validation accuracy of Figure~\ref{fig:val_acc}
terminates $2.6\%$ above the strongest baseline, and the policy entropy of
Figure~\ref{fig:bcp_entropy} rises monotonically from $0.23$ to $0.60$, the
only such trajectory in the comparison, so ENTR stabilizes training while
leaving exploration intact rather than obtaining stability by contracting the
policy. The downstream ordering matches, with ENTR at $37.34$ avg@1 ahead of
$34.94$ for KPop and $33.72$ for M2PO, within $1.21$ points of the synchronous
$38.55$.
On GSM8K, IcePop and KPop destabilize mid-training in
Figure~\ref{fig:gsm8k_score} while ENTR and M2PO train without disruption, and
they recover only $65.01$ and $70.51$ avg@4 against $95.28$ for M2PO and
$95.69$ for ENTR, whose pass@4 of $98.87$ exceeds the synchronous $98.49$.
Additional training dynamics for all four tasks are reported in
Appendix~\ref{app:dynamics}.

\subsection{Mathematical Reasoning}

DAPO-Math applies the most aggressive configuration we test,
$\Delta^{\mathrm{intra}}{=}8$ and $\Delta^{\mathrm{inter}}{=}10$.
In Figure~\ref{fig:dapo_score}, GRPO collapses irreversibly early
in training, IcePop and KPop plateau and then decline, and M2PO
saturates below ENTR, which leads throughout.
Averaged across AIME 2024 through 2026, ENTR reaches an avg@4
of $17.04$ against $16.29$ for M2PO, within $0.50$ points of
the synchronous $17.54$, and a pass@4 of $28.38$ that exceeds
both M2PO at $26.60$ and the synchronous $27.68$.

\paragraph{Mask behavior and variance control.}
Figures~\ref{fig:dapo_mask} and~\ref{fig:dapo_is_std} identify the cost at
which each keep rule obtains its stability, reporting the masked token
fraction $\rho_{\mathrm{mask}}$ together with the importance-ratio deviation.
ENTR terminates at $\rho_{\mathrm{mask}}{=}0.10\times10^{-4}$ and a deviation
of $2.62\times10^{-2}$, the smallest of the four on both axes, so it attains
the tightest variance control while sacrificing the fewest tokens. M2PO
discards roughly six times as many tokens yet ends with the largest deviation,
confirming that a larger $\rho_{\mathrm{mask}}$ does not imply tighter
control. ENTR combines a smaller importance-ratio deviation with a lower
masked fraction.

\subsection{Ablations and Sensitivity Analysis}

\paragraph{Ablating the two staleness sources.}
We vary one source of Eq.~\eqref{eq:staleness-decomp} at a time while holding
the other at its minimum, which isolates the degradation each contributes.
Figure~\ref{fig:intra_sweep} varies
$\Delta^{\mathrm{intra}}$ over $\{1,5,7,9\}$ at $\Delta^{\mathrm{inter}}{=}1$
and Figure~\ref{fig:inter_sweep} varies $\Delta^{\mathrm{inter}}$ over
$\{1,5,15,20,30\}$ at $\Delta^{\mathrm{intra}}{=}1$, both referenced against
the synchronous run.
ENTR remains stable across all configurations, while reward decreases
monotonically with either source of staleness.
The decline is slower for intra-trajectory staleness, consistent with
entropy normalization accommodating deviations at high-entropy
version-switch points.

\paragraph{Ablating exploration preservation and outlier suppression.}
On BrowseComp-Plus, we evaluate ENTR's preservation of
high-entropy exploration and suppression of low-entropy outliers.
Relative to fixed-magnitude masking, the former retains high-entropy
tokens rejected by the fixed threshold, while the latter filters
low-entropy tokens accepted by it.
As shown in Table~\ref{tab:bcp_token_ablation}, exploration preservation
and outlier suppression individually improve avg@1 by $2.55$ and
$2.20$ percentage points, respectively.
Combining both achieves the highest accuracy of $37.34\%$,
a $4.81$-point improvement over the fixed-magnitude baseline,
supporting the contribution of both interventions to held-out
performance.

\begin{table}[t]
\centering
\small
\renewcommand{\arraystretch}{1.12}
\setlength{\tabcolsep}{3pt}
\begin{tabular}{@{}lrr@{}}
\toprule
Variant & avg@1 (\%) & $\Delta$ (pp) \\
\midrule
Fixed-magnitude masking       & 32.53 & --    \\
\quad + High-entropy retention & 35.08 & +2.55 \\
\quad + Low-entropy filtering  & 34.73 & +2.20 \\
\midrule
\textbf{ENTR (both)} & \textbf{37.34} & \textbf{+4.81} \\
\bottomrule
\end{tabular}
\caption{Ablation of high-entropy retention and low-entropy
filtering on BrowseComp-Plus.}
\label{tab:bcp_token_ablation}
\end{table}

\paragraph{Masking budget.}
We calibrate every rule to the $0.07\%$ masked fraction of ENTR on a common
step-0 rollout batch and leave the threshold unconstrained thereafter, which
separates which tokens a rule removes from how many.
Table~\ref{tab:matched_budget} reports the outcome on multi-turn GSM8K.
Neither magnitude rule holds the budget, since a boundary constant in
$|\delta_t|$ ignores that the natural scale of the deviation grows with
entropy, and both drift above $17\%$ at accuracies below $71$. M2PO does hold
the budget at $0.24\%$ and still trails ENTR at $94.4$ against $95.7$, while
the tokens it removes have mean entropy $1.89$ against $0.11$. Holding the
budget is therefore necessary but not sufficient, and which tokens are removed
matters more than how many. Sensitivity to the remaining training
hyperparameters is reported in Appendix~\ref{app:sensitivity}.

\begin{table}[t]
\centering
\small
\setlength{\tabcolsep}{4pt}
\begin{tabular}{lccc}
\toprule
Method & $\rho_{\mathrm{mask}}$ (\%) & $\bar H_{\mathrm{masked}}$ & Acc.\ (\%) \\
\midrule
IcePop      & 0.07 $\to$ 19.16 & 3.47 & 65.3 \\
KPop        & 0.07 $\to$ 17.35 & 2.93 & 70.5 \\
M2PO        & 0.07 $\to$ 0.24  & 1.89 & 94.4 \\
\textit{ENTR (Ours)} & 0.07 $\to$ \textbf{\textit{0.12}}
 & \textbf{\textit{0.11}} & \textbf{\textit{95.7}} \\
\bottomrule
\end{tabular}
\caption{Matched-budget comparison of keep rules on multi-turn GSM8K.}
\label{tab:matched_budget}
\end{table}

\subsection{Training Efficiency}

Table~\ref{tab:efficiency} compares training efficiency under
identical hardware and parallelism. ENTR achieves a $2.6\times$
speedup over synchronous GRPO, with per-step runtime within $5\%$
of vanilla asynchronous GRPO. It also takes less time per step
than M2PO, at $514.84$\,s versus $602.78$\,s.
Together with the best overall asynchronous results reported in
Table~\ref{tab:main}, these findings show that ENTR improves
task performance while preserving the runtime efficiency of
asynchronous training.

\begin{table}[htbp]
\centering
\small
\setlength{\tabcolsep}{5pt}
\begin{tabular}{l c c c}
\toprule
Method & Throughput & s/step & Speedup \\
\midrule
GRPO (Sync)  & 82.56  & 1356.47 & $1.0\times$ \\
GRPO (Async) & 244.82 & 491.26  & $2.8\times$ \\
IcePop       & 228.54 & 503.56  & $2.7\times$ \\
KPop         & 217.30 & 510.12  & $2.7\times$ \\
M2PO         & 202.38 & 602.78  & $2.3\times$ \\
ENTR (Ours)  & 214.38 & 514.84  & $2.6\times$ \\
\bottomrule
\end{tabular}
\caption{Training efficiency on DAPO-Math.
Throughput is measured in tokens/s per GPU.
Speedup is computed from per-step wall-clock time
relative to synchronous GRPO.}
\label{tab:efficiency}
\end{table}

\section{Conclusion}
We introduce ENTR, an entropy-normalized token filter for stable
asynchronous LLM reinforcement learning. By adapting acceptance thresholds
to token entropy, ENTR suppresses low-entropy outliers while retaining
high-entropy updates. Across agentic and mathematical reasoning tasks,
it outperforms asynchronous baselines and approaches synchronous GRPO
performance with a $2.6\times$ speedup. Future work will explore longer
agentic horizons and more extreme asynchrony.

\section*{Limitations}
Our study focuses on asynchronous reinforcement learning for multi-turn tool
use and mathematical reasoning, and we leave other domains such as code
generation and multimodal reasoning to future work. We evaluate the keep rule
in isolation from complementary stabilization techniques such as replay
correction and value baselines, so their interaction with entropy
normalization is not characterized. Extending the analysis to substantially
longer agentic horizons and to larger training fleets is a natural next step.


\bibliography{custom}
\clearpage
\appendix

\section{An Entropy Envelope for Log-Importance Ratios}
\label{app:entropy-ratio}

\paragraph{Setup.}
At a fixed position $t$, let $\mu_t=\mathrm{softmax}(z)$ be the
behavior policy and $\pi_\theta=\mathrm{softmax}(z+\Delta)$ the
target policy. Define
$H_t=-\sum_w\mu_t(w)\log\mu_t(w)$,
$\mu_t^{\max}=\max_w\mu_t(w)$, and
$R_t=\max_w\Delta_w-\min_w\Delta_w$.
For $y\sim\mu_t$, write
$\delta(y)=\log(\pi_\theta(y)/\mu_t(y))$.

\begin{proposition}[Entropy envelope]
\label{prop:app-scaling}
For every finite logit difference $\Delta$,
\begin{equation}
\label{eq:app-law}
\begin{split}
\mathbb{E}_{\mu_t}[\delta^2]
&\le R_t(e^{R_t}-1)(1-\mu_t^{\max})\\
&\le R_t(e^{R_t}-1)H_t.
\end{split}
\end{equation}
\end{proposition}

\paragraph{Proof.}
The softmax identity
\begin{equation}
\label{eq:app-identity}
\delta(y)=\Delta_y-\log\mathbb{E}_{\mu_t}[e^\Delta]
\end{equation}
gives $\operatorname{Var}_{\mu_t}[\delta]
=\operatorname{Var}_{\mu_t}[\Delta]$ and
$\mathbb{E}_{\mu_t}[\delta]
=-D_{\mathrm{KL}}(\mu_t\Vert\pi_\theta)$. Hence,
\begin{equation}
\label{eq:app-second-moment}
\mathbb{E}_{\mu_t}[\delta^2]
=
\operatorname{Var}_{\mu_t}[\Delta]
+D_{\mathrm{KL}}(\mu_t\Vert\pi_\theta)^2.
\end{equation}
Choose $w^\star\in\arg\max_w\mu_t(w)$. Then
\begin{equation}
\label{eq:app-chain}
\begin{split}
\operatorname{Var}_{\mu_t}[\Delta]
&\le
\sum_{w\ne w^\star}
\mu_t(w)(\Delta_w-\Delta_{w^\star})^2\\
&\le R_t^2(1-\mu_t^{\max}).
\end{split}
\end{equation}

Applying $\log u\le u-1$ after subtracting
$\Delta_{w^\star}$ yields
\begin{equation}
\begin{split}
&D_{\mathrm{KL}}(\mu_t\Vert\pi_\theta)\\
&\quad\le
\mathbb{E}_{\mu_t}\!
\left[e^{\Delta-\Delta_{w^\star}}
-1-(\Delta-\Delta_{w^\star})\right]\\
&\quad\le
(1-\mu_t^{\max})(e^{R_t}-1-R_t).
\end{split}
\end{equation}
The last inequality follows because the integrand vanishes at
$w^\star$ and is at most $e^{R_t}-1-R_t$ whenever
$|\Delta_w-\Delta_{w^\star}|\le R_t$.

Moreover, $\log\mathbb{E}_{\mu_t}[e^\Delta]$ lies between
$\min_w\Delta_w$ and $\max_w\Delta_w$, so
$|\delta(y)|\le R_t$ and
$D_{\mathrm{KL}}(\mu_t\Vert\pi_\theta)\le R_t$.
Substituting into \eqref{eq:app-second-moment},
\begin{equation}
\begin{split}
\mathbb{E}_{\mu_t}[\delta^2]
&\le R_t^2(1-\mu_t^{\max})
+R_tD_{\mathrm{KL}}(\mu_t\Vert\pi_\theta)\\
&\le
\bigl[R_t^2+R_t(e^{R_t}-1-R_t)\bigr]
(1-\mu_t^{\max})\\
&=R_t(e^{R_t}-1)(1-\mu_t^{\max}).
\end{split}
\end{equation}
Finally,
$1-\mu_t^{\max}\le-\log\mu_t^{\max}\le H_t$
proves \eqref{eq:app-law}.

\paragraph{Implication for ENTR.}
At a fixed logit-gap range, the second-moment bound scales
linearly with entropy. For $\epsilon>0$,
\begin{equation}
\label{eq:app-normalized-bound}
\mathbb{E}_{\mu_t}
\left[\frac{\delta^2}{H_t+\epsilon}\right]
\le R_t(e^{R_t}-1).
\end{equation}
This motivates normalizing the squared log-ratio by entropy,
giving the ENTR acceptance rule
\begin{equation}
\frac{\delta_t^2}{H_t+\epsilon}\le\tau
\quad\Longleftrightarrow\quad
|\delta_t|\le\sqrt{\tau(H_t+\epsilon)}.
\end{equation}

\section{Low-Entropy Regime: Amplified Train--Inference Mismatch and the
Concave Arc}
\label{app:lowentropy-arc}

Appendix~\ref{app:entropy-ratio} bounds the deviation averaged over
$y\sim\mu_t$. We now characterize the conditional tail that appears when the
sampler draws a near-zero-probability token at a confident position. There the
local scale $a_t$ of Eq.~\eqref{eq:app-law} itself diverges, so the deviation
leaves the entropy envelope instead of following it.

\paragraph{Setup.}
Let $q_t\triangleq\pi_\theta(y_t\mid s_t)$ be the training-side probability of
the sampled token and let
\begin{equation}
\label{eq:app-mismatch}
\begin{split}
e_t&\triangleq\mu_t(y_t)-q_t,
\qquad\mathbb{E}[e_t\mid q_t]=0,\\
&\operatorname{Var}(e_t\mid q_t)=\lambda^2q_t(1-q_t),
\end{split}
\end{equation}
be the train--inference mismatch, where $\lambda$ is the discrepancy scale of
the two engines. Substituting into the definition of $\delta_t$ gives the exact
identity
\begin{equation}
\label{eq:app-b-exact}
\delta_t=-\log\!\Bigl(1+\frac{e_t}{q_t}\Bigr).
\end{equation}

\paragraph{Amplification by a near-zero denominator.}
The mismatch is absolute, but \eqref{eq:app-b-exact} measures it relative to
$q_t$. That relative discrepancy has standard deviation
\begin{equation}
\label{eq:app-rel}
\sqrt{\operatorname{Var}\bigl(e_t/q_t\mid q_t\bigr)}
=\lambda\sqrt{\frac{1-q_t}{q_t}}\xrightarrow[q_t\to0]{}\infty ,
\end{equation}
so an engine gap that is irrelevant at a high-entropy position is amplified
without bound as the model becomes confident. While the relative discrepancy
remains small, that is for $\lambda\ll\sqrt{q_t}$, expanding
\eqref{eq:app-b-exact} yields
\begin{equation}
\label{eq:app-varlaw}
\operatorname{Var}(\delta_t\mid q_t)
=\lambda^2\frac{1-q_t}{q_t}+O(\lambda^3).
\end{equation}
Outside that range the deviation is larger still, since $\delta_t$ grows
without bound as $\mu_t(y_t)\to0$, so \eqref{eq:app-varlaw} understates the
amplification rather than overstating it.

\begin{figure*}[t]
\centering
\includegraphics[width=\textwidth]{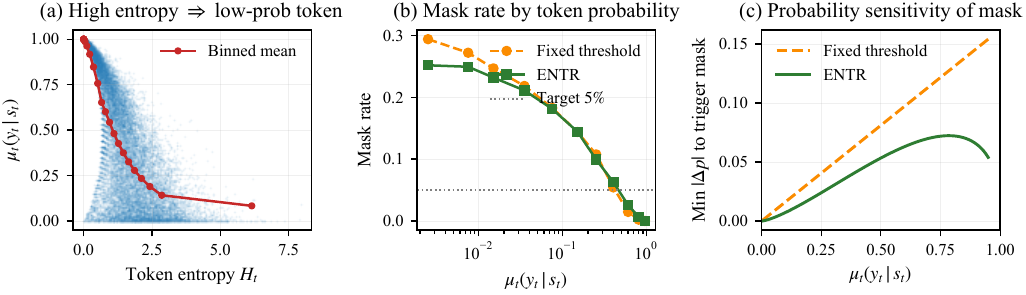}
\caption{The keep rule in probability space, on BrowseComp-Plus rollouts.
Sampled-token probability against token entropy, mask rate by sampled-token
probability, and the probability discrepancy needed to trigger masking.}
\label{fig:prob-fairness}

\end{figure*}

\paragraph{Binary-entropy envelope.}
A confident position places mass $1-q_t$ on its dominant token, and for a given
off-mode mass $q_t$ the entropy is smallest when the remainder sits on a single
alternative,
\begin{equation}
\label{eq:app-hbin}
H_t\;\ge\;H_{\mathrm{bin}}(q_t)=-q_t\log q_t-(1-q_t)\log(1-q_t),
\end{equation}
which is Eq.~\eqref{eq:hbin} of the main text and is independent of vocabulary
size. For small $q_t$ we have $H_{\mathrm{bin}}(q_t)=q_t(1-\log q_t)+O(q_t^2)$,
while \eqref{eq:app-varlaw} gives a deviation scale
$\bar\delta_t\triangleq\sqrt{\operatorname{Var}(\delta_t\mid q_t)}\propto
q_t^{-1/2}$. Eliminating $q_t$ between the two,
\begin{equation}
\label{eq:app-arc}
\bar\delta_t\approx\lambda\sqrt{\frac{1-\log q_t}{H_{\mathrm{bin}}(q_t)}},
\end{equation}
so the envelope opens without bound as $H_{\mathrm{bin}}(q_t)\to0$. This is the
concave arc of Figure~\ref{fig:joint}, and it reverses the high-entropy case,
where a large $|\delta_t|$ is licensed by a large $H_t$. A fixed threshold
keeps these tokens, whereas $b_{\mathrm{ENTR}}$ contracts toward its floor
$\sqrt{\tau\epsilon}$ here and removes them.

\paragraph{The keep rule in probability space.}
The amplification above is indexed by the probability of the sampled token
rather than by the entropy, since a confident position produces a large
deviation only when the sampler draws a non-dominant alternative.
Figure~\ref{fig:prob-fairness} therefore reads the keep rule against
$\mu_t(y_t\mid s_t)$ on BrowseComp-Plus rollouts, which equals $q_t$ up to the
mismatch $e_t$ of \eqref{eq:app-mismatch}. The sampled probability falls as
entropy rises, so the two coordinates run in opposite directions in the bulk
and the tokens analyzed here are the thin low-probability population that
survives at low entropy. Masking concentrates on improbable tokens under both
rules, and $b_{\mathrm{ENTR}}$ moves part of that budget toward confident
positions. Under a fixed threshold the discrepancy needed to trigger masking
grows linearly in the token probability, so mismatch on a confident token goes
unnoticed, whereas under $b_{\mathrm{ENTR}}$ the requirement saturates and the
rule stays sensitive there.

\section{Two-Sided Bounds on the Per-Token KL Divergence}
\label{app:kl_expansion}

This appendix proves Eq.~\eqref{eq:kl-sandwich}. Both distributions are
normalized, so $\mathbb{E}_{v\sim\mu_t}[\pi_\theta(v)/\mu_t(v)-1]=0$. Adding
this vanishing term to
$D_{\mathrm{KL}}(\mu_t\|\pi_\theta)=-\mathbb{E}_{v\sim\mu_t}[\delta_t(v)]$ and
substituting $\pi_\theta(v)/\mu_t(v)=e^{\delta_t(v)}$ gives Eq.~\eqref{eq:kl2},
\begin{equation}
\label{eq:app-kl-exact}
D_{\mathrm{KL}}(\mu_t\|\pi_\theta)
=\mathbb{E}_{v\sim\mu_t}\bigl[e^{\delta_t(v)}-1-\delta_t(v)\bigr],
\end{equation}
the $k_3$ estimator, which holds for every pair of distributions.

\begin{proposition}[Two-sided equivalence]
\label{prop:kl-sandwich}
Let $g(x)\triangleq(e^{x}-1-x)/x^{2}$ for $x\neq0$ and $g(0)\triangleq\tfrac12$.
Then $g$ is continuous and strictly increasing on $\mathbb{R}$.
For any $M>0$, if $|\delta_t(v)|\le M$ for every token $v$
with $\mu_t(v\mid s_t)>0$, then
\begin{equation}
\label{eq:app-kl-sandwich}
g(-M)\,\mathbb{E}\bigl[\delta_t^{2}\bigr]
\;\le\;D_{\mathrm{KL}}(\mu_t\|\pi_\theta)
\;\le\;g(M)\,\mathbb{E}\bigl[\delta_t^{2}\bigr].
\end{equation}
\end{proposition}

\begin{table*}[t]\centering\small
\setlength{\tabcolsep}{4pt}
\begin{tabular}{lcccc}
\toprule
 & Search-R1 & BrowseComp-Plus & DAPO-Math & GSM8K \\
\midrule
Policy model & Qwen3-8B & Qwen3-30B-A3B (MoE) & Qwen2.5-7B & Qwen2.5-7B \\
Train backend & Megatron & Megatron & Megatron & Megatron \\
Rollout engine & SGLang & SGLang & SGLang & SGLang \\
Task type & multi-turn search & multi-turn agentic & single-turn & multi-turn tool \\
Reward & rule-based EM & LLM judge & rule-based & rule-based \\
\midrule
Total nodes ($\times$8 H800) & 2 & 4 & 2 & 2 \\
Train nodes & 1 & 2 & 1 & 1 \\
Rollout nodes & 1 & 2 & 1 & 1 \\
Train GPUs / Rollout GPUs & 8 / 8 & 16 / 16 & 8 / 8 & 8 / 8 \\
\midrule
Actor TP/PP/CP/EP & 2/1/1/1 & 2/1/8/8 & 2/1/1/1 & 2/1/1/1 \\
Ref TP/PP/CP/EP & 2/1/1/1 & 2/1/4/8 & 2/1/1/1 & 2/1/1/1 \\
Rollout TP/EP & 2/1 & 8/1 & 2/1 & 2/1 \\
\midrule
Rollout samples $n$ & 4 & 8 & 8 & 8 \\
Global batch (prompts) & 128 & 16 & 128 & 256 \\
Prompt length & 4096 & 4096 & 2048 & 2048 \\
Response length & 8192 & 32768 & 12288 & 8192 \\
Max model length & 16384 & 40960 & 15376 & 12288 \\
Max turns & 8 & 100 & 100 & 100 \\
\midrule
ENTR $\tau$ & 1.6 & 1.0 & 1.6 & 1.6 \\
Entropy floor $\epsilon$ & $10^{-8}$ & $10^{-8}$ & $10^{-8}$ & $10^{-8}$ \\
\midrule
$\Delta^{\mathrm{intra}}$ & 2 & 5 & 8 & 5 \\
$\Delta^{\mathrm{inter}}$ & 9 & 1 & 10 & 13 \\
Staleness threshold & 9 & 5 & 20 & 10 \\
Param-sync interval & 1 & 1 & 2 & 4 \\
Partial rollout & yes & yes & yes & yes \\
Rollout mem.\ util. & 0.70 & 0.50 & 0.70 & 0.70 \\
Epochs & 5 & 5 & 12 & 12 \\
\bottomrule
\end{tabular}
\caption{Per-task training configuration. Every run is fully asynchronous with
disaggregated train and rollout pools.}
\label{tab:trainconfig}
\end{table*}

\paragraph{Proof.}
Differentiating gives
\begin{equation}
\begin{split}
g'(x)&=\frac{h(x)}{x^3},\\
h(x)&=(x-2)e^x+x+2.
\end{split}
\end{equation}
Here $h(0)=h'(0)=0$, and
\begin{equation}
\begin{split}
h'(x)&=(x-1)e^x+1,\\
h''(x)&=xe^x.
\end{split}
\end{equation}
Since $h''>0$ for $x>0$ and $h''<0$ for $x<0$,
the function $h'$ attains its minimum at the origin,
so $h'(x)>0$ for $x\ne0$ and $h$ is strictly increasing.
Together with $h(0)=0$, this gives $h(x)>0$ for $x>0$
and $h(x)<0$ for $x<0$. Hence $g'(x)>0$ for $x\ne0$.
Continuity at the origin follows from
$\lim_{x\to0}g(x)=\tfrac12$, establishing strict monotonicity
on $\mathbb{R}$.

Under this assumption, monotonicity bounds the integrand of
\eqref{eq:app-kl-exact} pointwise throughout the support of
$\mu_t$ as
\begin{equation}
g(-M)\delta_t^2
\le e^{\delta_t}-1-\delta_t
\le g(M)\delta_t^2.
\end{equation}
Taking expectations gives \eqref{eq:app-kl-sandwich}.
Writing the endpoints as
\begin{equation}
g(\mp M)=\frac{e^{\mp M}-1\pm M}{M^2}
\end{equation}
recovers Eq.~\eqref{eq:kl-sandwich}.

\paragraph{Reading the endpoints.}
Both endpoints tend to $\tfrac12$ as $M\to0$, giving the familiar
small-radius relation
$D_{\mathrm{KL}}\approx\tfrac12\mathbb{E}_{\mu_t}[\delta_t^2]$,
while the interval widens as $M$ grows.
For ENTR, $|\delta_t|\le b_{\mathrm{ENTR}}(H_t)$ holds for
retained tokens. The pointwise inequalities therefore apply
to the accepted region; integrating them with the keep mask
bounds that region's contribution to the KL integrand.
The rejected region remains part of the full KL divergence.

\paragraph{Second moment and dispersion.}
Eq.~\eqref{eq:app-identity} gives
$\mathbb{E}_{\mu_t}[\delta_t]=-D_{\mathrm{KL}}(\mu_t\|\pi_\theta)$, hence
\begin{equation}
\label{eq:app-kl-decomp}
\mathbb{E}\bigl[\delta_t^{2}\bigr]
=\operatorname{Var}_{\mu_t}[\delta_t]+D_{\mathrm{KL}}(\mu_t\|\pi_\theta)^{2}.
\end{equation}
This identity separates the second moment into the variance
of the log-ratio and the square of its mean.
Proposition~\ref{prop:app-scaling} bounds this second moment
under finite logit perturbations, providing a motivation
for entropy normalization.

\begin{figure*}[t]
\centering
\begin{subfigure}[t]{0.24\textwidth}
\centering
\includegraphics[width=\linewidth]{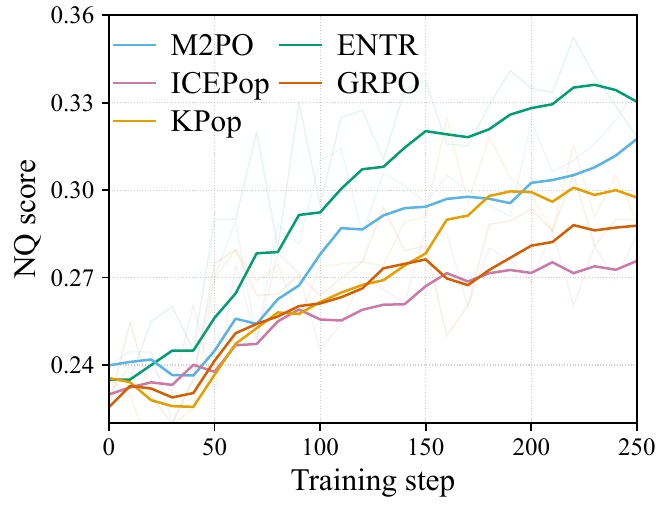}
\caption{NQ}
\label{fig:searchr1_nq}
\end{subfigure}
\hfill
\begin{subfigure}[t]{0.24\textwidth}
\centering
\includegraphics[width=\linewidth]{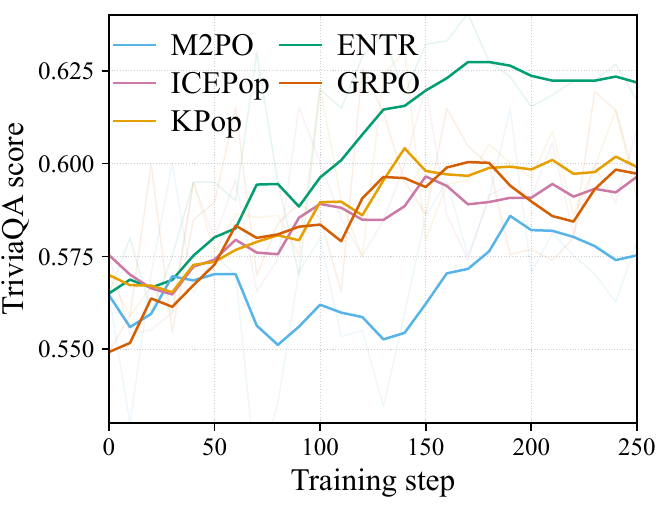}
\caption{TriviaQA}
\label{fig:searchr1_triviaqa}
\end{subfigure}
\hfill
\begin{subfigure}[t]{0.24\textwidth}
\centering
\includegraphics[width=\linewidth]{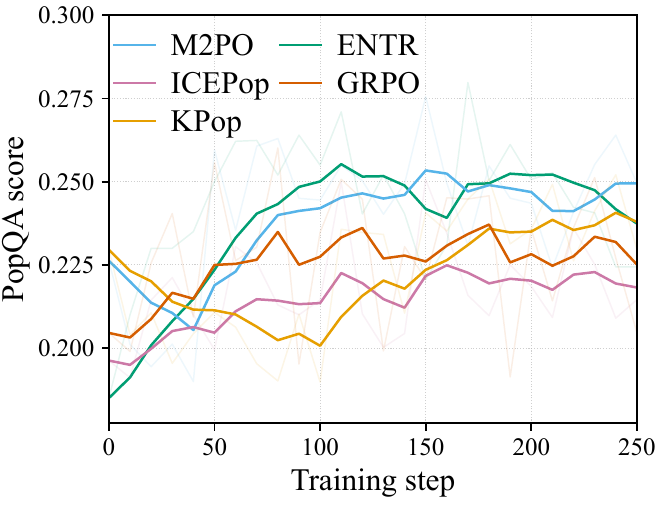}
\caption{PopQA}
\label{fig:searchr1_popqa}
\end{subfigure}
\hfill
\begin{subfigure}[t]{0.24\textwidth}
\centering
\includegraphics[width=\linewidth]{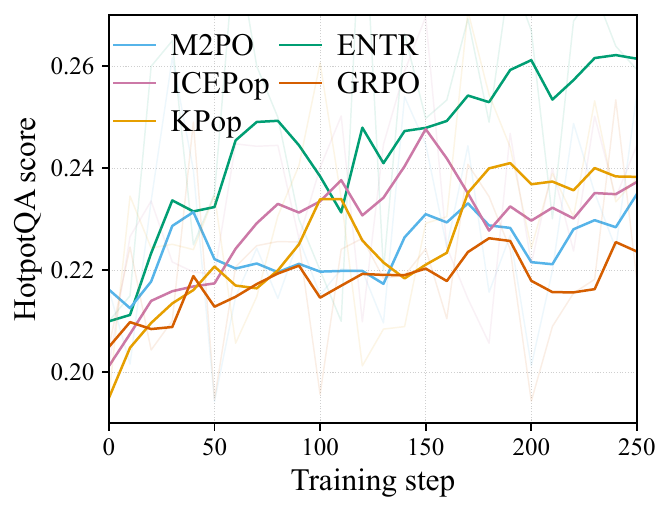}
\caption{HotpotQA}
\label{fig:searchr1_hotpotqa}
\end{subfigure}

\par\medskip

\begin{subfigure}[t]{0.24\textwidth}
\centering
\includegraphics[width=\linewidth]{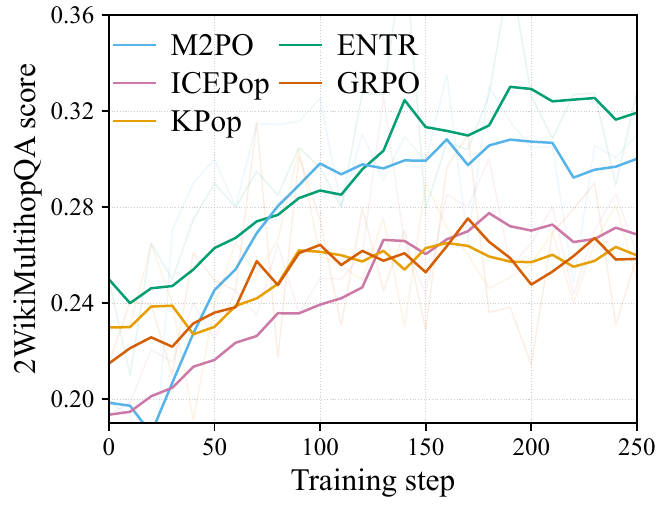}
\caption{2WikiMultihopQA}
\label{fig:searchr1_2wiki}
\end{subfigure}
\hspace{0.0133\textwidth}
\begin{subfigure}[t]{0.24\textwidth}
\centering
\includegraphics[width=\linewidth]{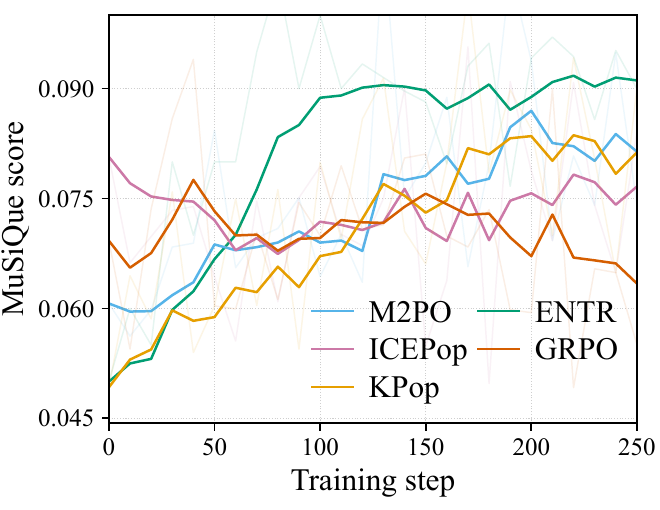}
\caption{MuSiQue}
\label{fig:searchr1_musique}
\end{subfigure}
\hspace{0.0133\textwidth}
\begin{subfigure}[t]{0.24\textwidth}
\centering
\includegraphics[width=\linewidth]{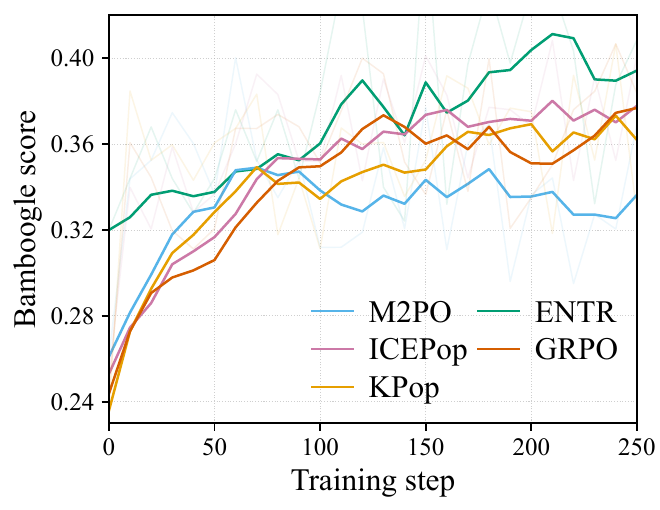}
\caption{Bamboogle}
\label{fig:searchr1_bamboogle}
\end{subfigure}

\caption{Validation dynamics on the seven Search-R1 evaluation sets with
Qwen3-8B under asynchronous training. NQ and HotpotQA are in-domain;
the remaining sets are out-of-domain. Solid curves show EMA-smoothed
scores with a span equivalent to $70$ training steps; faint curves show
the raw observations. All panels display the first $250$ training steps.
GRPO denotes the asynchronous baseline.}
\label{fig:searchr1_validation}
\end{figure*}

\section{Derivation of the ENTR Policy Gradient}
\label{app:grad}

Let $g_{i,t}$ be the clipped surrogate summed in Eq.~\eqref{eq:loss1}, so that
\begin{equation}
\label{eq:app-loss}
\mathcal{L}_{\mathrm{ENTR}}(\theta)
=-\,\mathbb{E}\Bigl[\frac{1}{\sum_{i}|o_i|}\sum_{i,t}M_{i,t}\,g_{i,t}\Bigr],
\end{equation}
and write
\begin{equation}
\label{eq:app-u}
u_{i,t}\triangleq\nabla_\theta\log\pi_\theta(o_{i,t}\mid s_{i,t}).
\end{equation}
The behavior policy $\mu_{i,t}$ that produced the rollout is fixed with respect
to $\theta$, and the keep mask $M_{i,t}=\mathbf{1}[S_{i,t}\le\tau]$ is piecewise
constant in $\theta$, so
\begin{equation}
\label{eq:app-dr}
\nabla_\theta r_{i,t}=r_{i,t}\,u_{i,t},\qquad\nabla_\theta M_{i,t}=0
\end{equation}
almost everywhere, and the mask acts as a stop-gradient gate. The gradient of
$g_{i,t}$ vanishes on the two flat clipped branches and elsewhere follows from
\eqref{eq:app-dr},
\begin{equation}
\label{eq:app-grad-cases}
\nabla_\theta g_{i,t}=
\left\{\begin{array}{@{}l@{\ }l@{}}
0 & A_{i,t}>0,\ r_{i,t}>1+\epsilon_{\mathrm{high}}\\[0.3ex]
0 & A_{i,t}<0,\ r_{i,t}<1-\epsilon_{\mathrm{low}}\\[0.3ex]
r_{i,t}A_{i,t}u_{i,t} & \text{otherwise}
\end{array}\right.
\end{equation}
Multiplying by $M_{i,t}$ and taking the expectation gives
Eq.~\eqref{eq:grad-exact}, in which $\kappa_{i,t}$ is the indicator of the third
branch. A token with $M_{i,t}=0$ contributes exactly zero, so the mask removes
the gradient of every token outside the trust region while leaving the gradient
of the retained tokens unchanged.

\section{Training Configuration}
\label{app:training-config}

All tasks use GRPO advantages with no KL penalty, asymmetric clipping bounds of
$0.20$ and $0.28$, a token-mean loss, no entropy bonus, and a constant learning
rate of $1\!\times\!10^{-6}$ under AdamW. ENTR adds only the token-level keep
rule $\delta_t^2/(H_t+\epsilon)\le\tau$ on top of GRPO, with the entropy floor
fixed at $\epsilon{=}1\!\times\!10^{-8}$ on every task and the rollout log-probabilities
serving as the proximal anchor. All runs offload parameters, gradients and
optimizer state in full and recompute activations. The complete per-task
configuration is summarized in Table~\ref{tab:trainconfig}.

\section{Additional Training Dynamics}
\label{app:dynamics}

\subsection{Search-R1}
\label{app:dynamics-searchr1}

Figure~\ref{fig:searchr1_validation} complements the aggregate training
reward with validation curves for each Search-R1 evaluation set.
ENTR finishes the displayed interval ahead of the asynchronous baselines
on six of the seven sets, including both in-domain sets. On NQ and
HotpotQA, its advantage emerges during the first half of training and
persists through step $250$.
Its gains also extend to TriviaQA, 2WikiMultihopQA, MuSiQue and Bamboogle,
indicating improvement beyond the training domains. On MuSiQue, ENTR
reaches a higher plateau by approximately step $100$ and maintains its
lead, while its Bamboogle score continues to improve later in training
despite local fluctuations.
PopQA presents a closer comparison, with ENTR and M2PO trading the lead
during training and M2PO ending slightly higher.
These curves show that ENTR's training reward gains are accompanied by
broad validation improvements, although the size and timing of the
advantage vary across datasets.

\subsection{BrowseComp-Plus}
\label{app:dynamics-bcp}

\begin{figure*}[t]
\centering
\begin{subfigure}[t]{0.32\textwidth}
\centering
\includegraphics[width=\linewidth]{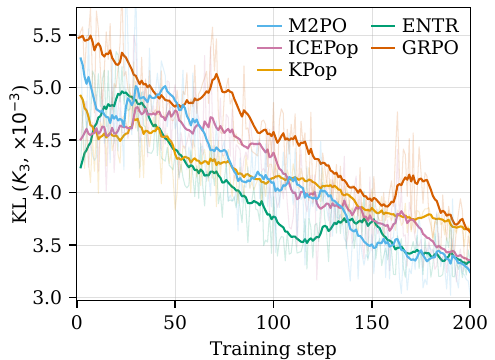}
\caption{KL to the rollout policy}
\label{fig:bcp_kl}
\end{subfigure}
\hfill
\begin{subfigure}[t]{0.32\textwidth}
\centering
\includegraphics[width=\linewidth]{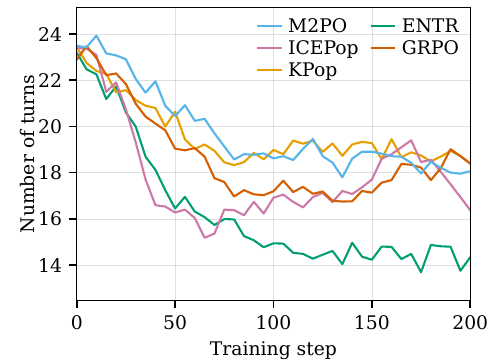}
\caption{Interaction turns}
\label{fig:bcp_num_turns}
\end{subfigure}
\hfill
\begin{subfigure}[t]{0.32\textwidth}
\centering
\includegraphics[width=\linewidth]{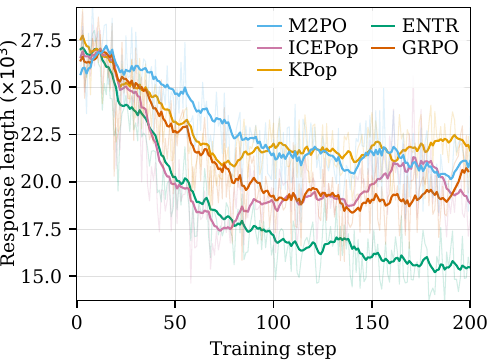}
\caption{Response length}
\label{fig:bcp_resp_len}
\end{subfigure}

\begin{subfigure}[t]{0.32\textwidth}
\centering
\includegraphics[width=\linewidth]{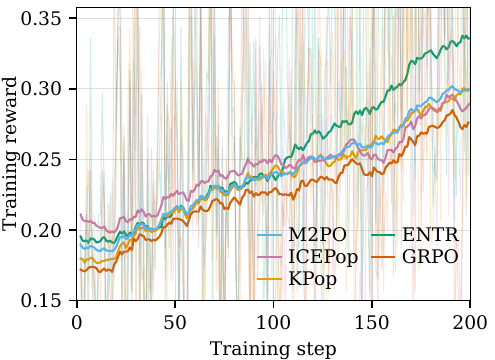}
\caption{Training reward}
\label{fig:bcp_score}
\end{subfigure}
\hfill
\begin{subfigure}[t]{0.32\textwidth}
\centering
\includegraphics[width=\linewidth]{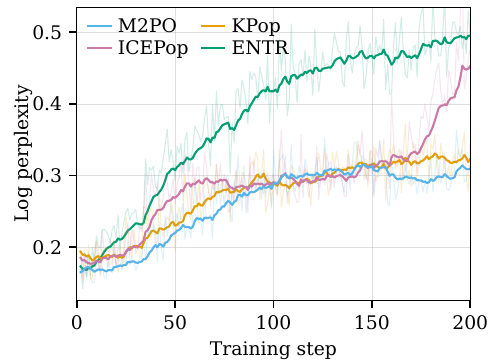}
\caption{Training-side log-perplexity}
\label{fig:bcp_log_ppl}
\end{subfigure}
\hfill
\begin{subfigure}[t]{0.32\textwidth}
\centering
\includegraphics[width=\linewidth]{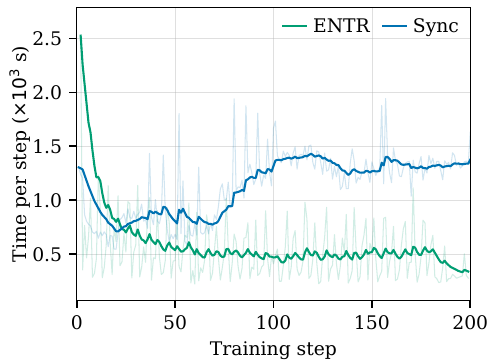}
\caption{Per-step wall-clock time}
\label{fig:bcp_time_per_step}
\end{subfigure}
\caption{Training dynamics on BrowseComp-Plus. Top, the $K_3$ estimate of the
KL to the rollout policy, the number of interaction turns, and the rollout
response length. Bottom, the training reward, the training-side log-perplexity
on the generated tokens, and the per-step wall-clock time. The log-perplexity
panel omits GRPO, and the timing panel compares ENTR against the synchronous
run.}
\label{fig:bcp_dynamics}
\end{figure*}

Figure~\ref{fig:bcp_dynamics} tracks six signals on BrowseComp-Plus. The KL to
the rollout policy in Figure~\ref{fig:bcp_kl} decays for every method, fastest
for ENTR and M2PO, which end near $3.3\times10^{-3}$, and slowest for KPop,
which ends at $3.8\times10^{-3}$. ENTR settles at the fewest turns and the
shortest responses, $14.3$ and $15.5\times10^3$ tokens in
Figures~\ref{fig:bcp_num_turns} and~\ref{fig:bcp_resp_len} against $18$ and
$21\times10^3$ for M2PO and KPop, yet reaches the highest reward in
Figure~\ref{fig:bcp_score} at $0.337$ with the baselines between $0.275$ and
$0.30$.

Log-perplexity in Figure~\ref{fig:bcp_log_ppl} measures how far the policy has
moved from the rollouts it scores and is not a quantity to be minimized. ENTR
rises smoothly to $0.50$ alongside its highest reward, while M2PO and KPop
plateau near $0.31$ and IcePop jumps to $0.45$ only after step $170$. M2PO
therefore matches ENTR on the KL with far less movement and a lower reward, so
a small gap to the rollout policy alone does not indicate effective learning.
In Figure~\ref{fig:bcp_time_per_step} the synchronous run climbs to
$1.4\times10^3$ s per step while ENTR holds near $0.5\times10^3$ s.

\begin{figure*}[t]
\centering
\includegraphics[width=0.99\textwidth]{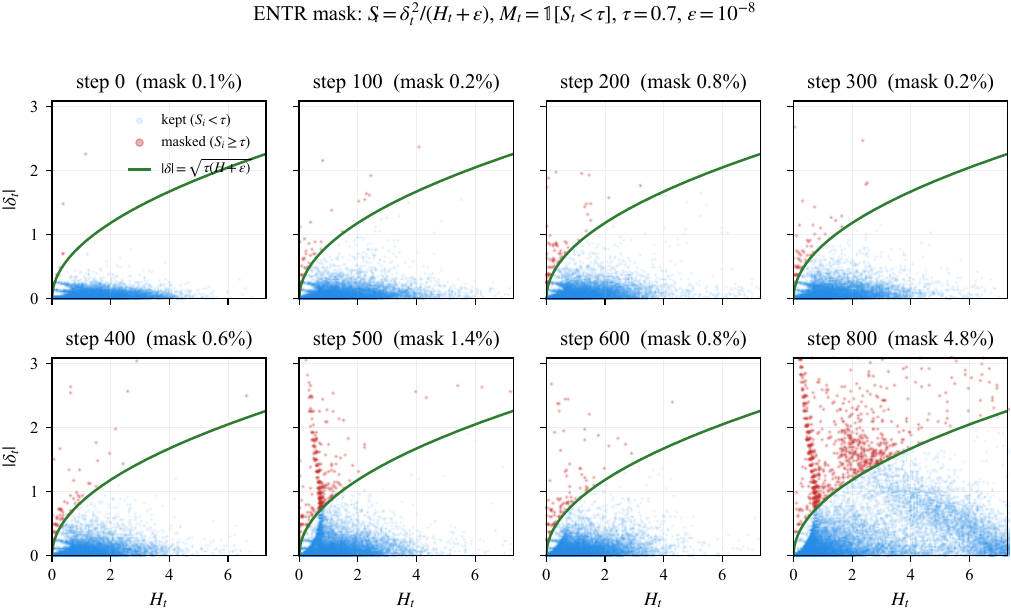}
\caption{Evolution of the ENTR keep mask in the $(H_t,|\delta_t|)$ plane on
BrowseComp-Plus. Each panel is the per-token joint distribution at one
training step, with kept tokens in blue, masked tokens in red and the boundary
$|\delta|=\sqrt{\tau(H_t+\epsilon)}$ in green. Titles give the masked fraction.
Shown at $\tau{=}0.7$, the most conservative setting of
Table~\ref{tab:sens-bcp-tau}, so that the masked population is visible; the
default $\tau{=}1.0$ gives the same geometry with a uniformly smaller masked
fraction.}
\label{fig:mask_evolution}
\end{figure*}

\paragraph{Mask evolution over training.}
Figure~\ref{fig:mask_evolution} shows the keep rule in the $(H_t,|\delta_t|)$
plane as training progresses. The spread of $|\delta_t|$ widens with $H_t$ and
the kept population fills the region beneath the boundary at every stage, so
the boundary follows the natural scale of the deviation rather than a fixed
cutoff. Masked tokens are dominated by low-entropy outliers carrying amplified
train-inference noise, while large deviations at high entropy stay inside the
boundary and are retained as exploration. The masked fraction is
self-regulating, below $1\%$ through most of training and rising to $4.8\%$
only at step $800$, when the accumulated off-policy gap inflates deviations
across the batch. The rule thus tightens only when harmful deviations emerge,
with no schedule and no staleness-specific tuning.

\subsection{Multi-turn GSM8K}
\label{app:dynamics-gsm8k}

\begin{figure*}[t]
\centering
\begin{subfigure}[t]{0.3\textwidth}
\centering
\includegraphics[width=\linewidth]{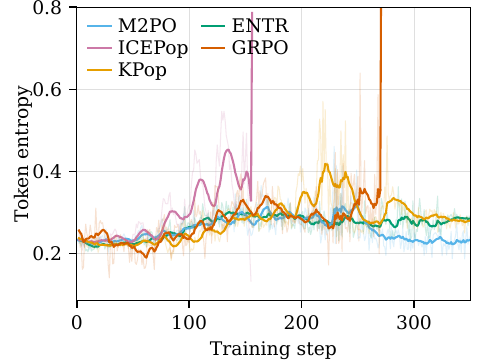}
\caption{Token entropy}
\label{fig:gsm8k_entropy}
\end{subfigure}
\hfill
\begin{subfigure}[t]{0.3\textwidth}
\centering
\includegraphics[width=\linewidth]{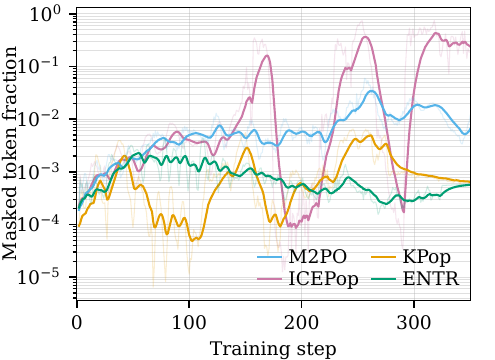}
\caption{Masked token fraction}
\label{fig:gsm8k_mask}
\end{subfigure}
\hfill
\begin{subfigure}[t]{0.3\textwidth}
\centering
\includegraphics[width=\linewidth]{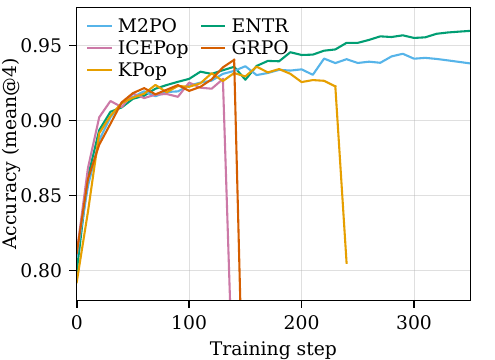}
\caption{Validation accuracy (avg@4)}
\label{fig:gsm8k_acc}
\end{subfigure}
\caption{Training dynamics on multi-turn GSM8K with Qwen2.5-7B. Token entropy,
the masked token fraction on a logarithmic axis, and validation accuracy at
avg@4. Each curve stops at its collapse onset.}
\label{fig:gsm8ks_dynamics}
\end{figure*}

Figures~\ref{fig:gsm8k_entropy} and~\ref{fig:gsm8k_mask} report entropy and
masking on multi-turn GSM8K. IcePop is the least stable, its entropy climbing
to $0.45$ before the run breaks at step $156$ and its masked fraction swinging
between $10^{-4}$ and $0.3$ across training. GRPO loses control of its entropy
later, at step $271$, and KPop peaks near $0.42$ around step $230$ before
settling back. ENTR and M2PO both hold entropy near $0.25$ throughout, but ENTR
masks an order of magnitude fewer tokens.

Accuracy in Figure~\ref{fig:gsm8k_acc} follows the same ordering. GRPO and
IcePop collapse once their entropy leaves its range, and KPop drops at step
$240$, with none of the three recovering. ENTR and M2PO are indistinguishable
until step $150$, after which ENTR continues to $0.96$ while M2PO flattens and
ends at $0.94$.

\subsection{DAPO-Math}
\label{app:dynamics-dapo}

\begin{figure*}[t]
\centering
\begin{subfigure}[t]{0.24\textwidth}
\centering
\includegraphics[width=\linewidth]{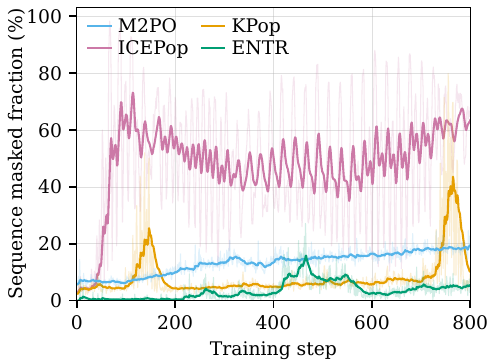}
\caption{Sequence masked fraction}
\label{fig:dapo_seq_mask}
\end{subfigure}
\hfill
\begin{subfigure}[t]{0.24\textwidth}
\centering
\includegraphics[width=\linewidth]{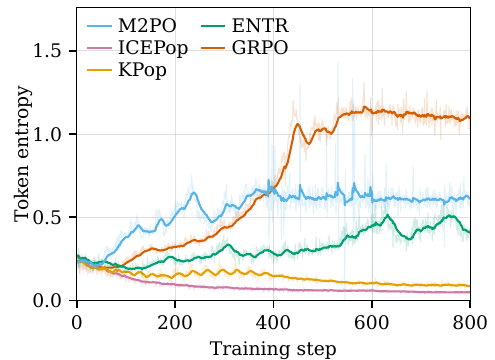}
\caption{Token entropy}
\label{fig:dapo_entropy}
\end{subfigure}
\hfill
\begin{subfigure}[t]{0.24\textwidth}
\centering
\includegraphics[width=\linewidth]{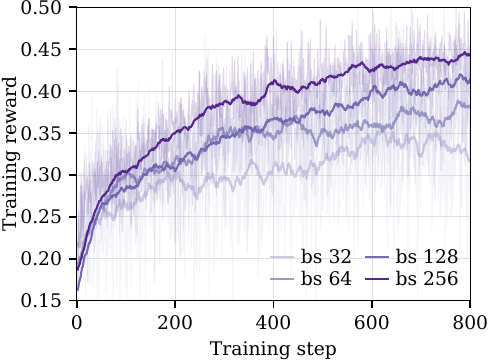}
\caption{Reward vs.\ batch size}
\label{fig:dapo_batch_score}
\end{subfigure}
\hfill
\begin{subfigure}[t]{0.24\textwidth}
\centering
\includegraphics[width=\linewidth]{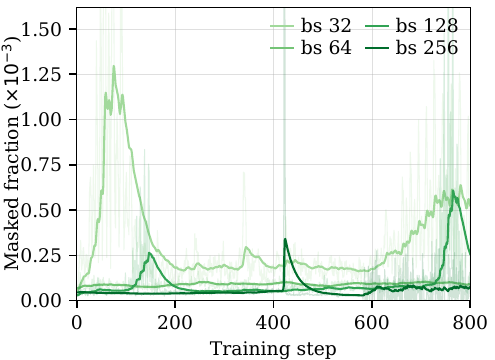}
\caption{Mask vs.\ batch size}
\label{fig:dapo_batch_mask}
\end{subfigure}
\caption{Training dynamics on DAPO-Math with Qwen2.5-7B. The sequence-level
masked fraction and the token entropy for the asynchronous methods, then the
sensitivity of ENTR to the global batch size over $32/64/128/256$ with darker
shades for larger batches, reported as training reward mapped from $[-1,1]$ to
$[0,1]$ and as the masked token fraction. Curves are EMA-smoothed and truncated
at $800$ steps.}
\label{fig:dapo_dynamics}
\end{figure*}

Figure~\ref{fig:dapo_seq_mask} reports the fraction of sequences containing at
least one masked token on DAPO-Math. IcePop sits between $40\%$ and $70\%$
throughout, KPop stays near $5\%$ with transient spikes to $25\%$ and $43\%$,
and M2PO drifts up from $6\%$ to $19\%$, whereas ENTR remains below $8\%$ apart
from a single spike at step $465$. Entropy in Figure~\ref{fig:dapo_entropy}
separates the two failure modes. GRPO rises to $1.1$ and stays there, while
IcePop and KPop fall below $0.1$ through over-suppression. ENTR sustains a
gradual rise to about $0.45$ and M2PO settles higher at $0.62$, so the point is
not how many tokens a rule removes but which ones.

Figures~\ref{fig:dapo_batch_score} and~\ref{fig:dapo_batch_mask} sweep the
global batch over $32/64/128/256$. Training reward improves monotonically with
batch size, from $0.32$ to $0.44$, with diminishing returns beyond $128$. The
masked fraction stays below $1.3\times10^{-3}$ across the sweep and shows no
systematic ordering in batch size, so larger batches help through variance
reduction rather than by altering the trust region, matching
Table~\ref{tab:sens-dapo-gbs}.

\section{Hyperparameter Sensitivity}
\label{app:sensitivity}

We vary the keep threshold $\tau$ and the global batch size while holding the
remaining settings fixed. All AIME results use avg@4.

\paragraph{Keep threshold.}
On BrowseComp-Plus the spread over $\tau\in[0.7,1.6]$ is $1.4$ points in
accuracy and $2.4$ points in reward, with $\tau{=}0.7$ weakest on both.

\begin{table}[htbp]\centering
\begin{tabular}{lccc}
\toprule
 & $\tau{=}0.7$ & $\tau{=}1.0$ & $\tau{=}1.6$ \\
\midrule
Accuracy & 35.97 & 37.34 & 37.15 \\
Reward & 29.87 & 32.25 & 30.62 \\
\bottomrule
\end{tabular}
\caption{Sensitivity to the ENTR keep threshold $\tau$ on BrowseComp-Plus
(Qwen3-30B-A3B); other settings as in Table~\ref{tab:trainconfig}.}
\label{tab:sens-bcp-tau}
\end{table}

On DAPO-Math the spread is at most $2.0$ points and $\tau{=}0.7$ is again
weakest on every set. The defaults of Table~\ref{tab:trainconfig}, $\tau{=}1.0$
on BrowseComp-Plus and $\tau{=}1.6$ on DAPO-Math, are either best or within
$0.5$ points of best.

\begin{table}[htbp]\centering
\begin{tabular}{lccc}
\toprule
$\tau$ & $0.7$ & $1.0$ & $1.6$ \\
\midrule
AIME-2024 (avg@4) & 19.67 & 19.85 & 20.03 \\
AIME-2025 (avg@4) & 14.33 & 16.12 & 15.64 \\
AIME-2026 (avg@4) & 13.49 & 15.21 & 15.46 \\
\bottomrule
\end{tabular}
\caption{Sensitivity to $\tau$ on DAPO-Math (Qwen2.5-7B); other settings as in
Table~\ref{tab:trainconfig}.}
\label{tab:sens-dapo-tau}
\end{table}

\paragraph{Global batch size.}
Raising the batch from $32$ to $128$ prompts gains between $2.4$ and $3.1$
points on each AIME set, while $256$ moves the result by less than one point in
either direction, so the benefit saturates near $128$.

\begin{table}[htbp]\centering
\small
\setlength{\tabcolsep}{3pt}
\begin{tabular}{lcccc}
\toprule
Batch size & $32$ & $64$ & $128$ & $256$ \\
\midrule
AIME24 avg@4 & 16.97 & 18.26 & 20.03 & 19.87 \\
AIME25 avg@4 & 13.23 & 14.19 & 15.64 & 15.58 \\
AIME26 avg@4 & 12.94 & 14.47 & 15.46 & 16.32 \\
\bottomrule
\end{tabular}
\caption{Sensitivity to global batch size (prompts) on DAPO-Math (Qwen2.5-7B);
$\tau{=}1.6$, other settings as in Table~\ref{tab:trainconfig}.}
\label{tab:sens-dapo-gbs}
\end{table}

\end{document}